\documentclass[11pt, a4paper]{article}
\usepackage[T1]{fontenc}
\usepackage{amsmath, amssymb, mathtools}
\usepackage{amsthm} % before cleveref, so that \cref names theorems defined in main.tex
\usepackage{stix2}
\usepackage[varqu, varl, scaled=0.95]{inconsolata}
\usepackage[english]{babel}
\usepackage{microtype}
\usepackage[a4paper, hmargin=2.75cm, vmargin=3cm, headheight=30pt, headsep=0.75cm, footskip=1.2cm, heightrounded]{geometry}

\usepackage{enumitem}
\setlist{itemsep=2pt, parsep=0pt, topsep=0pt}


\usepackage{xcolor}
\definecolor{accent}{rgb}{0.09, 0.5, 0.99}
\colorlet{quiet}{black!55}

\makeatletter
\g@addto@macro\normalsize{%
  \setlength{\abovedisplayskip}{10pt plus 2pt}%
  \setlength{\belowdisplayskip}{10pt plus 2pt}%
  \setlength{\abovedisplayshortskip}{\abovedisplayskip}%
  \setlength{\belowdisplayshortskip}{\belowdisplayskip}%
}
\makeatother

\usepackage{titlesec}
\titleformat{\section}{\normalfont\Large\bfseries}{\thesection}{1em}{}[{\suppressfloats[t]}]
\titleformat{\subsection}{\normalfont\large\bfseries}{\thesubsection}{1em}{}
\titleformat{\subsubsection}{\normalfont\normalsize\bfseries}{\thesubsubsection}{1em}{}
\titlespacing*{\section}{0pt}{24pt plus 4pt minus 2pt}{9pt plus 1pt}
\titlespacing*{\subsection}{0pt}{17pt plus 3pt minus 2pt}{5pt plus 1pt}
\titlespacing*{\subsubsection}{0pt}{13pt plus 2pt minus 2pt}{3pt plus 1pt}

\newcommand{\sectionpage}{\clearpage}
\AddToHook{cmd/section/before}{\sectionpage}
\AddToHook{cmd/appendix/after}{\clearpage{\LARGE\bfseries Appendix\par}\gdef\sectionpage{\gdef\sectionpage{\clearpage}}%
  \addtocontents{toc}{\protect\setcounter{tocdepth}{1}}}

\usepackage{fancyhdr}
\makeatletter
\newcommand{\subtitle}[1]{\def\@subtitle{#1}}
\newcommand{\headnote}[1]{\def\@headnote{#1}}
\newcommand{\contact}[1]{\def\@contact{#1}}
\newcommand{\runningtitle}[1]{\def\@runningtitle{#1}}
\subtitle{}
\headnote{}
\contact{}
\def\@runningtitle{\@title}
\fancypagestyle{first}{%
  \fancyhf{}%
  \fancyhead[L]{\small\@headnote}%
}
\makeatother

\AddToHook{cmd/tableofcontents/before}{\begingroup\setlength{\parskip}{0.3\baselineskip}\hypersetup{linkcolor=black}}
\AddToHook{cmd/tableofcontents/after}{\endgroup}

\usepackage{graphicx}
\graphicspath{{figures/}}
\usepackage{booktabs}
\usepackage{caption}
\usepackage[font=footnotesize, labelfont=bf]{subcaption}
\usepackage[hang, flushmargin]{footmisc}
\usepackage[numbers, sort&compress]{natbib}
\newcommand{\doi}[1]{\href{https://doi.org/#1}{doi:#1}}

\usepackage[bookmarksnumbered, bookmarksopen, bookmarksdepth=3]{hyperref}
\hypersetup{colorlinks=true, allcolors=accent}
\makeatletter
\AtBeginDocument{\hypersetup{pdftitle={\@title\ifx\@subtitle\@empty\else: \@subtitle\fi}, pdfauthor={\@author}}}
\makeatother
\usepackage{xurl} % URLs and file names break anywhere
\newcommand{\file}{\nolinkurl} % \file{code/make_figures.py}: a file name, underscores as typed
\usepackage[capitalize, noabbrev]{cleveref}
\crefname{equation}{Eq.}{Eqs.}
\Crefname{equation}{Equation}{Equations}
\newcommand{\sectionsign}[1]{%
  \crefformat{#1}{\S##2##1##3}%
  \Crefformat{#1}{Section~##2##1##3}%
  \crefrangeformat{#1}{\S##3##1##4--##5##2##6}%
  \Crefrangeformat{#1}{Sections~##3##1##4--##5##2##6}%
  \crefmultiformat{#1}{\S##2##1##3}{ and~\S##2##1##3}{, \S##2##1##3}{ and~\S##2##1##3}%
  \Crefmultiformat{#1}{Sections~##2##1##3}{ and~##2##1##3}{, ##2##1##3}{ and~##2##1##3}%
}
\sectionsign{section}
\sectionsign{subsection}
\sectionsign{subsubsection}
\AddToHook{cmd/appendix/after}{\crefalias{section}{appendix}\crefalias{subsection}{subappendix}} % \cref writes "Appendix A", "Appendix A.1"

\makeatletter
\renewcommand{\maketitle}{%
  \thispagestyle{first}%
  \begingroup
  \setlength{\parskip}{0pt}\raggedright%
  \vspace*{0.8cm}%
  {\fontsize{22}{27}\selectfont\bfseries\@title\par}%
  \ifx\@subtitle\@empty\else
  \vspace{0.35cm}%
  {\fontsize{14}{19}\selectfont\@subtitle\par}%
  \fi
  \vspace{0.9cm}%
  {\large\@author\par}%
  \vspace{0.1cm}%
  {\small\ttfamily\color{quiet}\@contact\par}%
  \endgroup
}
\renewenvironment{abstract}{%
  \par\vspace{1cm}%
  {\large\bfseries\abstractname\par}%
  \noindent\ignorespaces%
}{%
  \par
}
\makeatother

\usepackage{array} % for the ragged-right p columns of the tables

\makeatletter\def\NAT@spacechar{~}\patchcmd{\NAT@test}{\else\NAT@nm}{\else\mbox{\NAT@nm}}{}{\PackageError{main}{NAT@test patch failed}{}}\makeatother % \citet keeps the author names and the bracket on one line
\setcitestyle{notesep={,~}} % and the number with its note, as in [64, Prop.~2]
\newcommand{\seedrange}[2]{\unskip\,{\scriptsize\color{black!60}[$#1$,\,$#2$]}} % a thin space from the value and inside the brackets
\newcommand{\load}[1]{\ell_{\mkern-3mu #1}} % the load of an expert, its subscript set close to the STIX Two ell
\newcommand{\vx}{\mathbf{x}}
\newcommand{\vc}{\mathbf{c}}
\newcommand{\vw}{\mathbf{w}}
\newcommand{\Wr}{\mathbf{W}_{\mathrm{r}}}
\newcommand{\R}{\mathbb{R}}

\title{Muon Makes the First Mixture-of-Experts Layer Harder to Balance}
\runningtitle{Muon Makes the First MoE Layer Harder to Balance}
\author{Marcos Hernanz\\[2pt]{\normalsize Escuela Politécnica Superior, Universidad Autónoma de Madrid}}
\contact{marcos@hernanz.io}
\headnote{\textbf{Technical report}}

\begin{document}
\medmuskip=4mu plus 2mu minus 1mu\relax
\hypersetup{pdfauthor={Marcos Hernanz}}

\maketitle
\begin{NoHyper}{\setlength{\footnotemargin}{0pt}\renewcommand{\thefootnote}{}\footnotetext{This paper is a short version of the author's bachelor's thesis at Universidad Autónoma de Madrid, supervised by Daniel Hernández-Lobato.}}\end{NoHyper}

\begin{abstract}
  Many open mixture-of-experts (MoE) language models train with Muon instead of AdamW\@. Most balance the load with a bias on each expert's routing score, adjusted after every step. Their recipes disagree on whether Muon should also train the router. In models with 55M and 108M active parameters, we measure how the optimizer changes the span of these biases, the largest minus the smallest in a layer. The span grows when all tokens favor the same experts, so it shows how hard a layer is to balance. At learning rates tuned on shorter runs, Muon raised the span of the first MoE layer at every seed, roughly doubling it on average, while the other layers changed much less. The rise came from Muon on the rest of the network, whichever optimizer trained the router. In those runs, the inputs to the first router shared a direction, and soon after the warmup its busiest expert averaged four to seven times its fair share. Over the first 3,000 steps, subtracting the batch mean from these inputs removed this overload while barely changing the loss. On the smaller model, halving Muon's learning rate removed most of the rise and slightly improved the loss. Moving only the router to AdamW made the loads shift more during the warmup and slightly raised the loss, so at this scale the results favor keeping the router on Muon.
\end{abstract}

\tableofcontents

\section{Introduction}
\label{sec:introduction}

Most of the leading open language models of 2026 are mixture-of-experts (MoE) models~\citep{deepseek2026v4, kimi2026k3, zai2026glm53, qiu2026qwennext, xiaomi2026mimov26}. An MoE layer has many feed-forward networks, the experts, and a router that sends each token to a few of them~\citep{shazeer2017outrageously, lepikhin2021gshard, fedus2022switch}. Left alone, the router tends to send most tokens to a few experts, which then train on more data and attract still more tokens~\citep{shazeer2017outrageously}. Earlier models such as GShard and the Switch Transformer countered this with an auxiliary balance loss that penalizes uneven loads~\citep{shazeer2017outrageously, lepikhin2021gshard, fedus2022switch}. DeepSeek-V3 switched to auxiliary-loss-free balancing~\citep{wang2024lossfree, deepseek2024v3}, and most recent open models have followed, among them DeepSeek-V4, GLM-5.3, and MiMo-V2.6~\citep{deepseek2026v4, zai2026glm53, xiaomi2026mimoprocard}, some keeping a small balance loss as well. Each expert has a bias that is added to its score when the router selects experts. After every step, the bias of an expert that received more than its fair share of the tokens moves down by a fixed amount, and the bias of an expert that received less moves up. We refer to this balancer as the bias.

\begin{figure}[!b]
  \centering
  \includegraphics{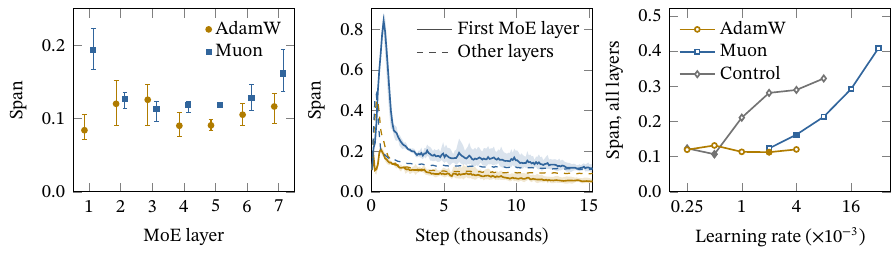}
  \caption{The span, the largest minus the smallest bias of a layer, under AdamW and Muon. Left, the mean span of each MoE layer in the main comparison (small model, 1B tokens, seeds 0 to~2), with its range over the seeds. Center, the span of the first MoE layer and the mean span of the other six layers in the same runs, colored as in the left panel, in bins of 100 steps, with the range over the seeds shaded for the first MoE layer. Right, the span averaged over the layers in the tuning runs (0.5B tokens, seed~0) against the peak learning rate, with filled marks at the rates selected by the tuning. Where marks overlap, at $2.5 \times 10^{-4}$ and $2 \times 10^{-3}$, the spans differ by 0.011 or less. The control is Muon without the orthogonalization (\cref{sec:partitions}).}
  \label{fig:main}
\end{figure}

\looseness=-1 Most of the leading open MoE models now train their weight matrices with Muon~\citep{jordan2024muon, liu2025moonlight} instead of AdamW~\citep{loshchilov2019decoupled}. Muon orthogonalizes the update of each weight matrix, which sets all the singular values of the update to one. The recipes differ on whether Muon trains the router (\cref{tab:recipes}). DeepSeek\nobreakdash-V4, Kimi~K3, and GLM-5.3 do not exclude it from Muon, while Qwen3.8-Flash-Next, MiMo-V2.6, and IQuest-Q1 do. Only the report of Qwen3.8-Flash-Next gives reasons, among them that Muon made the router unstable early in training with no significant gain later~\citep{qiu2026qwennext}. No model report shows data that compare the options, and the studies that compare optimizers in MoE models train without the bias or do not measure the bias itself~\citep{wang2026depth, lau2026symmetry, srivastava2026lossfree, liu2025moonlight, xie2026spectralsphere, vaidya2026compass, malik2026skewadam, miasnikov2026monarch, jin2026silent}. In one of them, which trains a model with about 60B parameters balanced by the bias, the loads of the first MoE layer were far more uneven early in training under Muon than under AdamW, and more uneven still with the router kept on AdamW, in one run per arm and with the cause left open~\citep{jin2026silent}.

The bias pushes the loads toward their fair shares whatever the router prefers, so the loads say little about the router. We measure instead how hard the bias works, by its span, the largest minus the smallest bias in a layer (\cref{sec:span}). The span stays near zero when the scores of the router spread the tokens evenly on their own, and it grows with any preference for some experts that all tokens share. When the span is large, the bias decides much of the routing, and the loads stay uneven while it catches up with changes in the scores.

We study how the optimizer changes the span in two MoE language models trained on FineWeb-Edu~\citep{penedo2024fineweb}, a small one with 55M and a large one with 108M active parameters. In each run we choose which optimizer, AdamW or Muon, trains each group of parameters, so that the router, the experts, and the other parts of the network can be moved to Muon separately. The backbone is every weight matrix except the routers, the embedding table, and the output head. Runs with the same seed start from the same weights and see the same batches, so within each experiment we compare runs with the same seed, at three seeds. The code and the runs are public (\cref{app:reproduction}).

At learning rates tuned for each optimizer on shorter runs, Muon raised the span of the first MoE layer at every seed, from 0.084 to~0.194 on average (\cref{fig:main}), and it did so again at all nine seeds of three later experiments (\cref{sec:experiments}). We had expected the opposite. Under Muon the update of an expert has the same size however many tokens the expert receives, which should weaken the feedback in which busy experts become busier. The span of the first MoE layer followed the optimizer of the backbone and changed little with the optimizer of the router (\cref{sec:backbone}). With the backbone on Muon, the inputs of the router of the first MoE layer shared a direction, with a mean cosine similarity of about 0.4 against 0.07 under AdamW on the small model. At the tuned rates, the mean of these inputs gave every token the same preference among the experts. The bias took hundreds of steps to cancel this preference, and meanwhile, with the bias alone at its usual rate, the busiest expert of the layer received on average four to seven times its fair share. In runs that repeated the first 3,000 steps of both models, subtracting the batch mean from the input of this router removed the overload and changed the loss by less than 0.003 (\cref{sec:direction,sec:robustness}).

None of the settings that put only part of the backbone on Muon reproduced the shared direction, yet two of them raised the span of the first MoE layer more than the whole backbone did, so a large span can arise without it (\cref{sec:direction}). On the small model, half the tuned rate of Muon trained slightly better at 1B tokens and removed about three quarters of the rise, although the shared direction remained (\cref{sec:rate}). A bias four times faster halved the rise and removed most of the overload after the warmup, and the rise persisted with an added balance loss and on the large model, where the first MoE layer was still overloaded at the end of training (\cref{sec:robustness}).

With the backbone on Muon, moving the router to AdamW made the loads shift more during the warmup, on both model sizes and at every learning rate of a stress test, and at the tuned rates it raised the loss slightly at every seed (\cref{sec:router}). In the large-model experiment, loss spikes occurred only in the runs of Muon with the router on AdamW, although the tuning of the large model had spikes in its run of Muon at twice the selected learning rate. For models of our size, these results favor keeping the router on Muon when the backbone is on Muon, and they point to the first MoE layer as the one most affected by the optimizer. Our models have about 50 to~2,000 times fewer active parameters than those whose recipes motivate the question (\cref{sec:limitations}).

\section{Related Work}
\label{sec:related}

\begin{table}[!b]
  \centering
  \caption{The optimizer of the router and the balancer in recent open MoE models. In the router column, Muon means that the report, or the earlier report it cites, does not exclude the router from Muon. The balancer is stated in the report, in the earlier report it cites, or in the released configuration~\citep{zai2026glm53, qwen2026flashnextcard, xiaomi2026mimoprocard}.}
  \label{tab:recipes}
  \small
  \begin{tabular}{@{}lll>{\raggedright\arraybackslash}p{6.5cm}@{}}
    \toprule
    & & & What the report says \\
    Model & Router & Balancer & about the router or the balance \\
    \midrule
    DeepSeek-V4 & Muon & bias, balance loss & loss spikes tied to outliers in the MoE layers, which routing appears to exacerbate \\ \addlinespace[2pt]
    Kimi~K3 & Muon & bias set from the scores & a fixed bias step adapts slowly or oscillates \\ \addlinespace[2pt]
    GLM-5.3 & Muon & bias, balance loss & cites GLM-5, which follows the training setting of GLM-4.5, whose Muon excludes only embeddings, bias vectors, and normalization gains \\ \addlinespace[2pt]
    Qwen3.8-Flash-Next & AdamW & balance loss & Muon made the router unstable early, with no significant gain later \\ \addlinespace[2pt]
    MiMo-V2.6 & AdamW & bias & frozen in reinforcement learning after it unbalanced the loads \\ \addlinespace[2pt]
    IQuest-Q1 & Adam & not stated & a dense first layer, to mitigate imbalance in the earliest layer \\
    \bottomrule
  \end{tabular}
\end{table}

\Cref{tab:recipes} lists the optimizer of the router and the balancer in recent open MoE models. DeepSeek-V4 names only the embedding table, the output head, the normalization gains, and a few small parameters as trained with AdamW, the later DeepSeek-V4.1-Flash trains the weight matrices of the linear layers of its backbone with Muon, and Kimi~K3 lists no exception to Muon among its weight matrices~\citep{deepseek2026v4, deepseek2026v41flash, kimi2026k3}. GLM-5.3 shares the base model of GLM-5.2 and cites the GLM-5 report, which follows the training setting of GLM-4.5. That setting excludes only the embeddings, the bias vectors, and the normalization gains from Muon~\citep{zai2026glm53, glm2026glm5, glm2025glm45}. We therefore take the routers of DeepSeek-V4, Kimi~K3, and GLM-5.3 to be on Muon, although no report states which optimizer trains the router. For GLM-5.3 this assumes that GLM-5.2, which changed the architecture of GLM-5~\citep{zai2026glm52}, kept the training setting of GLM-5. The report of Qwen3.8-Flash-Next, which keeps its router on AdamW, suggests that each output of the router, the score of one expert, is largely independent of the others, leaving no shared linear structure for the orthogonalization to exploit~\citep{qiu2026qwennext}. MiMo-V2.6 moved most of its weight matrices from AdamW to Muown, a variant of Muon~\citep{lion2026muown}, in mid-training and kept the router on AdamW without a stated reason, and it froze the router during reinforcement learning after the busiest expert of a layer reached 16 times its fair share~\citep{xiaomi2026mimov26}. IQuest-Q1 trains its router with Adam, again without a stated reason~\citep{yang2026iquest}.

The study closest to ours is that of \citet{wang2026depth}, who show that a full-rank update of Muon moves every row of the router by the same distance. In a 500M-parameter MoE trained on math text without any balancer, with one seed and both optimizers at a learning rate of $10^{-3}$, the usage of the experts became unbalanced under AdamW and stayed more even under Muon over the first 1,000 steps. The exception was the first layer in their runs without depth scaling, where both were about equally uneven. Their experiments switch the optimizer of the whole model at once, while we switch the router and parts of the rest separately. Their companion paper describes a direction shared by the hidden states of trained MoE models and argues that a balance loss teaches the router to suppress it, a suppression they observe in gpt-oss-20b and DeepSeek-V2-Lite, trained with AdamW and a balance loss, but not in Moonlight and Trinity~Mini, trained with Muon and without the loss~\citep{wang2026myth}. \citet{xie2026spectralsphere} switch the optimizer of the whole model in an MoE balanced by the bias, at a learning rate tuned for AdamW and with their own scaling of the update of Muon, and, in one run per optimizer, find the maximal violation of the model as a whole~\citep{wang2024lossfree} higher under AdamW than under Muon until about the last tenth of training, where the two are about equal. The maximal violation is our overload (\cref{sec:measurements}) minus one and measures the imbalance that the bias leaves. In a blog post, \citet{jin2026silent} train an MoE with about 60B parameters, 1B of them active, and 768 routed experts, of which a sigmoid router selects 8, balanced by the bias at an update rate of 0.001. Like our models, it has a shared expert and one dense layer before the first MoE layer. With the backbone on Muon, the maximal violation of the first MoE layer was far higher early in training than under AdamW, and higher still with the router on AdamW, under which some experts were briefly selected by every token. The seventh MoE layer showed the same order early in training. Their comparison has one run per arm and leaves the cause open. \citet{li2026spectralcap} finds that in the first MoE layer of a model trained with Muon and the bias, one direction carries 99.5\% of the variance of the logits plus the bias and one expert is among the four selected for every token. In the last of its six layers, one expert is among the four selected for 82\% of the tokens. Its MoE runs use one seed, and none of them trains the router with AdamW\@.

Other work changes the update of only part of the network. \citet{lau2026symmetry} derive updates of the router from its symmetries. In one run each, without a balancing bias and with the attention and the experts on Muon, these updates and a plain Muon-style update of the router reach a lower loss and more even loads than a router on AdamW, which shares the learning rate of all but one of them. In an essay whose runs keep the attention and the experts on Muon and balance the loads by the bias or one of two variants of it, a router on Muon at $2 \times 10^{-2}$ is better conditioned than one on Adam at $3 \times 10^{-3}$, at nearly the same loss, in one plotted run each whose balancer is not stated~\citep{srivastava2026lossfree}. The gain of a concurrent router-only update rule depends on the base optimizer and on the renormalization of the mixing weights~\citep{anon2026sqrt}. ExpertMuon-Compass~\citep{vaidya2026compass}, concurrent work on FineWeb-Edu with small MoE models balanced by the bias, changes only the step sizes of the expert matrices, updates the router alike in the two arms of its routing analysis, and so attributes the routing differences to the updates of the experts. A memory-saving study~\citep{malik2026skewadam} compares AdamW with Muon in an MoE trained with a balance loss, keeping the router on Adam in both, at $10^{-3}$ with Muon and $3 \times 10^{-4}$ with AdamW\@. In its single Muon run, at untuned rates, the balance loss rose to 22\% above its floor over the last 1,000 of 10,000 steps. A study of structured Muon updates trains a 1B-parameter MoE balanced by the bias with AdamW and with Muon at one shared rate, keeping the router on AdamW in both, and reports only the validation loss of one run each~\citep{miasnikov2026monarch}. None of the studies above compares the work of the bias in runs that differ only in the optimizer.

\citet{wang2024lossfree}, who introduced the bias, kept an update rate of 0.001 after trying 0.0001 and 0.01. \citet{han2025lossfree} prove that, with the scores of the router held fixed and a small enough update rate, this update brings the token count of every expert within $E - 1$ of its fair share for $E$ experts. They extend the analysis to scores drawn independently from fixed distributions, replacing this guarantee with a regret bound and the fixed-rate sign update with an update proportional to the difference between each load and its fair share, at a rate that falls in inverse proportion to the number of updates so far. In neither case do the scores respond to the bias. Trinity~Large damps the update of the bias because, its authors hypothesize, the undamped update oscillates near balance~\citep{singh2026trinity}, Kimi~K3 sets the bias of each step from quantiles of the router scores of the previous batch, each taken relative to the routing cutoff of its token~\citep{kimi2026k3}, and $\phi$-balancing instead weights a balance loss by the gradient of a convex potential at a moving average of the mean router probabilities~\citep{chen2026phibalancing}. Two concurrent works read the bias as an integral feedback controller~\citep{jin2026idbalancing, anon2026feedback}, and \citet{jin2026idbalancing} read the quantile rule of Kimi~K3 as a generalized proportional one.

Several reports single out the lowest MoE layers or the start of training. The routing of the first layer of OLMoE settles much later than that of the others~\citep{muennighoff2025olmoe}, and in preliminary runs of Sigma-MoE-Tiny, which selects one expert per token, the bias, combined with a balance loss, grew in the lower layers until it decided the routing~\citep{hu2025sigmamoe}. Motif~3, trained with Muon, gave its earliest MoE layers a larger balance-loss coefficient because they were more prone to imbalance early in training, and their routing still collapsed occasionally~\citep{motif2026motif3}. In preliminary runs of Laguna~M.1, which trains with a balance loss, Muon without the learning-rate scaling of Moonlight and with standard weight decay made the experts collapse from about 450B tokens on, layer by layer through the first three MoE layers, until training diverged~\citep{poolside2026laguna}. In early runs of Trinity~Large, which trains with Muon, the experts collapsed, after which \citet{singh2026trinity} made six changes at once, among them the damped update of the bias, a sequence-wise balance loss, and six dense layers instead of three, with no ablation to tell which one helped. In the 60B-parameter model of \citet{jin2026silent} trained with AdamW, the weight decay shrank the weights of the routed experts of the lowest MoE layers, which received little gradient, until masking these experts barely changed the model. Under Muon these weights did not collapse, but masking the experts still changed the model little. A language-modeling loss on the output of the first MoE layer, applied early in training, kept these experts in use. DeepSeekMoE makes its first layer dense because the loads of that layer balanced more slowly~\citep{dai2024deepseekmoe}, and IQuest-Q1 does so to mitigate imbalance in the earliest layer~\citep{yang2026iquest}. DeepSeek-V4 routes its first three MoE layers by a hash of the token id~\citep{deepseek2026v4, roller2021hash}. A concurrent submission traces routing collapse under AdamW to the failure of the router to correct perturbations that coincide with an early reorganization of the hidden representations~\citep{anon2026stuck}. In another, which trains OLMoE models from scratch, the first layer was hard for the bias to balance and became the worst-balanced layer after a change of the data mixture, in the one run per balancer whose layers it reports~\citep{anon2026feedback}. In Nemotron~3 Ultra, trained with AdamW, the bias, and a balance loss, the busiest expert of the first MoE layer received 4.8 times its fair share early in pretraining and about 12 times by 12T tokens, while the median layer stayed near 1.2~\citep{nvidia2026ultra, nvidia2026super}. The report found this imbalance while looking for the cause of a later divergence and links it to training instability, and it does not say whether these loads were measured on training or validation tokens. Apart from \citet{jin2026silent}, none of these works compares AdamW with Muon.

\section{Background and Measurements}
\label{sec:background}

\subsection{The MoE Layer and the Bias}
\label{sec:moe}

Our MoE layer follows DeepSeekMoE~\citep{dai2024deepseekmoe} with the routing of DeepSeek-V3~\citep{deepseek2024v3}. Let $\vx \in \R^{d}$ be the normalized hidden state of a token at the input of the layer. The layer has one shared expert $\operatorname{FFN}_{\mathrm{s}}$ and $E$ routed experts $\operatorname{FFN}_1, \dots, \operatorname{FFN}_E$, each a feed-forward network (FFN), of which it selects $k$ for each token. The router $\Wr \in \R^{E \times d}$ gives expert $i$ the score $s_i = \sigma(\vw_i^{\top} \vx)$, where $\vw_i$ is the $i$-th row of $\Wr$ and $\sigma$ the logistic sigmoid. The layer computes
\begin{equation}
  \mathcal{T}(\vx) = \operatorname{Top}_k \Bigl( \bigl\{ s_i + b_i \bigr\}_{i=1}^{E} \Bigr), \qquad
  \mathbf{y} = \operatorname{FFN}_{\mathrm{s}}(\vx) + \sum_{i \in \mathcal{T}(\vx)} \frac{s_i}{\sum_{j \in \mathcal{T}(\vx)} s_j} \operatorname{FFN}_i(\vx),
  \label{eq:moe}
\end{equation}
where $b_i$ is the bias of expert $i$. The bias decides which experts are selected but does not enter the mixing weights, and, without a balance loss, the router receives gradient only through the mixing weights of the selected experts. The layer has no capacity limit, so every token reaches its $k$ experts. In a batch of $N$ tokens, the load $\load{i}$ of expert $i$ is the fraction of the $Nk$ selections that go to it, and its fair share is $1/E$. After every optimizer step, auxiliary-loss-free balancing~\citep{wang2024lossfree} updates
\begin{equation}
  b_i \leftarrow b_i + u \operatorname{sign}\bigl( 1/E - \load{i} \bigr), \qquad i = 1, \dots, E,
  \label{eq:bias}
\end{equation}
with a fixed update rate $u$. The bias starts at zero and receives no gradient.

\subsection{The Span and the Shared Direction}
\label{sec:span}

We call $\max_i b_i - \min_i b_i$ the span of a layer. Adding the same constant to every bias leaves the selection unchanged, so only the differences between biases matter, and the span is the largest of them. If the router gives one expert a score 0.1 higher than another for every token, the two receive equal loads only when the bias of the first is about 0.1 below that of the second, so the span is then about 0.1 or more. The span has a ceiling. When $b_i - b_j \geq 1$, expert $i$ ranks above expert $j$ for every token, since the scores lie in $(0, 1)$. Then $\load{i} \geq \load{j}$, and \cref{eq:bias} cannot widen the gap. A difference between two biases therefore grows only while it is below 1, by at most $2u$ per step, and the span never exceeds $1 + 2u$.

Following \citet{wang2026myth}, let $\vc$ be the mean of the router inputs $\vx_1, \dots, \vx_N$ of a batch. The logit of expert $i$ for token $n$ splits as
\begin{equation}
  \vw_i^{\top} \vx_n = \vw_i^{\top} \vc + \vw_i^{\top} (\vx_n - \vc),
  \label{eq:offset}
\end{equation}
and the first term, the offset of expert $i$, is the same for every token. When the inputs share a direction, $\vc$ is long compared with the deviations $\vx_n - \vc$. Differences between the offsets of the experts then become a preference that every token shares and that the bias has to cancel. This preference grows with the length of the mean input and with the spread of the rows of the router along it.

\subsection{Muon}
\label{sec:muon}

Following \citet{liu2025moonlight}, our update of Muon for a weight matrix $\mathbf{W} \in \R^{p \times q}$ with gradient $\mathbf{G}_t$ at step $t$ is
\begin{equation}
  \mathbf{M}_t = \mu \mathbf{M}_{t-1} + \mathbf{G}_t, \qquad
  \mathbf{W}_t = \mathbf{W}_{t-1} - \eta_t \bigl( 0.2 \sqrt{\max(p, q)} \operatorname{Ortho}(\mu \mathbf{M}_t + \mathbf{G}_t) + \lambda \mathbf{W}_{t-1} \bigr),
  \label{eq:muon}
\end{equation}
where $\mathbf{M}_t$ is the momentum with coefficient $\mu$, $\eta_t$ is the learning rate, $\lambda$ the weight decay, and $\operatorname{Ortho}(\mathbf{A}) = \mathbf{U} \mathbf{V}^{\top}$ for the reduced singular value decomposition $\mathbf{A} = \mathbf{U} \mathbf{\Sigma} \mathbf{V}^{\top}$. Muon approximates $\operatorname{Ortho}$ with a few Newton--Schulz steps~\citep{jordan2024muon}. The factor $0.2 \sqrt{\max(p, q)}$ gives the update a root mean square close to that of a typical update of AdamW\@. Under Muon, every matrix of every expert receives, apart from the weight decay, an update with roughly the same spectral norm, $0.2\, \eta_t \sqrt{\max(p, q)}$, whatever the load of the expert. Under AdamW, the gradient of an expert with a low load is summed over few tokens and fluctuates more from step to step relative to its mean. AdamW divides a running average of the gradient by the root mean square of its recent values, so such an expert can be expected to move less. We therefore expected Muon to weaken the feedback in which busy experts become busier, and to lower the span.

\subsection{Measurements}
\label{sec:measurements}

Every 10 steps, a run logs metrics for every MoE layer on the batch of that step. The \emph{input cosine} is the mean cosine similarity between the router inputs of two distinct tokens,
\begin{equation*}
  \rho = \frac{1}{N(N-1)} \sum_{n \neq m} \hat{\vx}_n^{\top} \hat{\vx}_m = \frac{\lVert \sum_{n} \hat{\vx}_n \rVert^2 - N}{N(N-1)}, \qquad \hat{\vx}_n = \frac{\vx_n}{\lVert \vx_n \rVert},
\end{equation*}
near 0 when the inputs point in all directions and 1 when they share one, a measure of anisotropy~\citep{ethayarajh2019contextual}. The \emph{score offset} ${\max_i \bar{s}_i - \min_i \bar{s}_i}$, with $\bar{s}_i$ the mean score of expert $i$ over the batch, tracks the spread of the offsets of \cref{eq:offset}, the largest minus the smallest offset, in units of the score. The \emph{overload} $E \max_i \load{i}$ is the load of the busiest expert as a multiple of its fair share, from 1 when the loads are even to~$E/k$ when the same $k$ experts take every token. With the $E = 32$ and $k = 4$ of our models (\cref{sec:models}), this ceiling is 8. The \emph{load churn} is the total variation distance $\sum_i\, \lvert \load{i} - \ell'_{\mkern-3mu i} \rvert / 2$ between the loads $\load{i}$ at a logged step and the loads $\ell'_{\mkern-3mu i}$ at the logged step before it, related to the routing fluctuation of StableMoE~\citep{dai2022stablemoe}.

The \emph{common share} $\kappa$ is the fraction of a step $\Delta \Wr$ of the router that acts along the shared direction,
\begin{equation*}
  \kappa = \frac{\lVert \Delta \Wr \, \mathbf{e} \rVert}{\lVert \Delta \Wr \rVert_{\mathrm{F}}},
\end{equation*}
where $\mathbf{e}$ is the unit vector along $\smash[b]{\sum_n \hat{\vx}_n}$, the direction of the mean normalized input. An update with $E$ equal singular values has $\kappa \leq 1/\sqrt{E}$, which is about 0.18 for $E = 32$. The singular values of an update of Muon are nearly equal, so its common share stays near this bound, while that of AdamW has no such bound. The \emph{update--load correlation} is the Pearson correlation across the experts of a layer between the number of tokens that an expert receives in a step and the Frobenius norm of the change that the step makes to its weight matrices, or in a second version the largest of their spectral norms. The \emph{usage perplexity} $\exp(-\sum_i \load{i} \log \load{i})$ is the effective number of experts in use~\citep{wang2026depth}. The \emph{specialization} is the Jensen--Shannon divergence among the loads measured on four text domains at the last checkpoint, which OLMoE compares expert by expert~\citep{muennighoff2025olmoe}.

A metric becomes one number per run by averaging over the logged steps of a window, in the first MoE layer or over all MoE layers. The integrated value, which we report unless we name another window, averages from the end of the 500-step warmup to the end of the run. The early value averages over the $0.05\, S$ steps after the warmup of a run of $S$ steps, which are steps 501 to~1,262 with 1B tokens and 501 to~2,788 with 3B tokens, and the warmup value averages over steps 1 to~500, with the load churn of the first logged step taken as zero. We call the early value of the overload of the first MoE layer the \emph{early overload}, and the warmup value of the load churn over all MoE layers the \emph{warmup churn}.

\section{Experimental Setup}
\label{sec:setup}

\subsection{Models and Data}
\label{sec:models}

The models are decoder-only Transformers~\citep{vaswani2017attention, radford2019language} with RMSNorm~\citep{zhang2019rmsnorm}, rotary position embeddings~\citep{su2024roformer}, normalized queries and keys~\citep{henry2020qknorm, dehghani2023vit22b}, gated attention~\citep{qiu2025gated}, and SwiGLU feed-forward networks~\citep{shazeer2020glu}. Block 0 has a dense FFN, and every later block has the MoE layer of \cref{sec:moe} with $E = 32$ routed experts, of which $k = 4$ are selected. We call the MoE layer of block~1 the \emph{first MoE layer} and its router the \emph{first router}. The small model has 8 blocks of width 512 and 55M active parameters out of 303M, and the large model 12 blocks of width 768 and 108M active out of 496M, with the embedding table and the output head counted only in the totals (\cref{tab:architecture} in \cref{app:configuration}). Both train on sequences of 1,024 tokens in batches of 64, so an expert with its fair share receives 8,192 tokens per step. The bias follows \cref{eq:bias} with $u = 0.001$, the rate of DeepSeek-V3 and DeepSeek-V4~\citep{deepseek2024v3, deepseek2026v4}, and keeps this rate until the last step, while DeepSeek-V3 stopped the updates for its last 500B tokens. The bias is the only balancer except where we add a balance loss.

The training data are the sample-10BT subset of FineWeb-Edu~\citep{penedo2024fineweb}, split into tokens by byte-level byte-pair encoding~\citep{sennrich2016bpe, radford2019language} with a vocabulary of 16,384. A permutation fixed by the seed orders the training sequences. The validation loss is the mean cross-entropy on the same 2.1M held-out tokens in every run. The specialization uses 2M tokens from each of four domains, held-out FineWeb-Edu, OpenWebMath~\citep{paster2024openwebmath}, codeparrot-clean-valid~\citep{codeparrot2022clean}, and the English Wikipedia~\citep{wikimedia2023wikipedia}.

\subsection{Partitions and Optimizers}
\label{sec:partitions}

We divide the parameters into six groups, which are the routers, the routed experts, the shared experts, the attention, the dense FFN of block~0, and a last group with the embedding table, the output head, and the gains of the normalization layers. AdamW trains the last group in every run, as in the reports that use Muon~\citep{jordan2024muon, liu2025moonlight, deepseek2026v4}. A \emph{partition} assigns AdamW or Muon to each of the other five groups, and the four groups other than the routers form the backbone. \Cref{tab:partitions} lists the partitions we use. The first three, which we call the \emph{recipe partitions}, are the choices of current recipes. DeepSeek-V3 trains every weight matrix with AdamW~\citep{deepseek2024v3}, DeepSeek-V4 and Moonlight train every weight matrix with Muon except the embedding table, the output head, and a few small ones in DeepSeek-V4~\citep{deepseek2026v4, liu2025moonlight}, and Qwen3.8-Flash-Next keeps the router on AdamW~\citep{qiu2026qwennext}. We call the second and the third the two \emph{Muon partitions}. In every partition, each optimizer uses its own tuned rate (\cref{sec:rates}), except where an experiment scales the rates.

\begin{table}[t]
  \centering
  \caption{The partitions and the control. The five groups are the routers, the routed experts, the shared experts, the attention, and the dense FFN of block~0. A partition trains with AdamW every group that it does not put on Muon, and the embedding table, the output head, and the normalization gains always with~AdamW.}
  \label{tab:partitions}
  \small
  \begin{tabular}{@{}ll>{\raggedright\arraybackslash}p{4.3cm}@{}}
    \toprule
    Partition & Groups trained with Muon & Experiments \\
    \midrule
    AdamW & none & all \\
    Muon & all five & all \\
    Muon, router on AdamW & all but the router & stress test, ablation, balance~loss, large model \\
    AdamW, router on Muon & the router & ablation \\
    Experts on Muon & the routed experts & ablation \\
    Attention on Muon & the attention & ablation \\
    Dense FFN on Muon & the dense FFN of block~0 & ablation \\
    Attention, dense FFN on Muon & the attention and the dense FFN & ablation \\
    Control & all five, without the orthogonalization & tuning \\
    \bottomrule
  \end{tabular}
\end{table}

Muon follows \cref{eq:muon} with $\mu = 0.95$, five Newton--Schulz steps, and $\lambda = 0.1$, as in Moonlight~\citep{liu2025moonlight}. It treats the router of each layer as one matrix and, as DeepSeek-V4 does, each of the three matrices of every routed expert as a separate matrix. AdamW uses $\beta_1 = 0.9$, $\beta_2 = 0.95$, and $\lambda = 0.1$, as in DeepSeek-V3 and DeepSeek-V4~\citep{deepseek2024v3, deepseek2026v4}. The \emph{control} keeps the momentum of Muon and the size of its update but drops the orthogonalization. It rescales the Nesterov momentum $\mu \mathbf{M}_t + \mathbf{G}_t$ of each matrix to the Frobenius norm $\sqrt{\min(p, q)}$ of an orthogonal matrix of full rank and then applies the factor and the weight decay of \cref{eq:muon}. The updates of all experts therefore have the same size, as under Muon, while the singular values of each update stay as unequal as those of the momentum.

\subsection{Learning Rates}
\label{sec:rates}

Since the best learning rates of AdamW and Muon differ, we tune the peak rate of each optimizer separately~\citep{wen2025fantastic, semenov2025benchmarking}, on the partition that trains all five groups with it. The tuning uses the small model, seed~0, and 0.5B tokens, with five rates for AdamW and Muon and six for the control, each twice the one before, and selects the rate with the lowest final validation loss. It selects $2 \times 10^{-3}$ for AdamW, $4 \times 10^{-3}$ for Muon, and $5 \times 10^{-4}$ for the control, none at the end of its grid, and the losses of Muon at $2 \times 10^{-3}$ and $4 \times 10^{-3}$ differ by 0.003. AdamW is tuned first, and in every run of Muon or of the control, AdamW trains the embedding table, the output head, and the gains at the rate tuned for AdamW\@. The small model uses these rates with 1B tokens, twice the tuning length. The large model is tuned again with 1B tokens at 0.125 to~1 times the rates of the small model, and this tuning selects $10^{-3}$ for AdamW and $2 \times 10^{-3}$ for Muon. The large model then trains on 3B tokens, so we multiply both rates by 0.7, about $3^{-0.32}$, following the fall of the best rate with the number of tokens that \citet{bjorck2025horizons} fit for AdamW on models of 760M parameters or more and \citet{ren2026hypersphere} for Muon constrained to a hypersphere at 208M parameters. The rate of each optimizer rises linearly over 500 steps and then decays along a cosine~\citep{loshchilov2017sgdr} to a tenth of its peak, and each optimizer clips the gradient of its parameters at norm 1.0 (\cref{app:configuration}).

\subsection{Experiments and Comparisons}
\label{sec:experiments}

An \emph{arm} of an experiment is a partition with its learning rates and balancer, and \cref{app:program} lists every arm and run in \cref{tab:runs}. The \emph{main comparison} trains AdamW and Muon on the small model with 1B tokens and seeds 0 to~2. The \emph{stress test} trains the three recipe partitions with 0.5B tokens and seeds 0 to~2 at 1, 2, 4, 8, and 16 times their tuned rates, and Muon at 4 times with a \emph{fast bias} that moves by $4u$ per step. As in the stress test of Qwen3.8-Flash-Next~\citep{qiu2026qwennext}, the rates do not decay, and they stay constant after the warmup. The stress test clips the gradient at twice the median norm after the warmup in a calibration run, which trains the same partition at its tuned rates. The \emph{ablation}, with 1B tokens and seeds 3 to~5, trains the four combinations of AdamW and Muon for the router and the backbone, Muon at half its tuned rate, the four partitions with part of the backbone on Muon (\cref{tab:partitions}), and AdamW and Muon without the bias. Two experiments with 1B tokens vary the balancer. The \emph{bias-speed} experiment trains AdamW and Muon with the usual bias and with the fast bias, at seeds 6 to~8. The \emph{balance-loss} experiment adds the sequence-wise balance loss of DeepSeek-V3~\citep{deepseek2024v3} with weight $\alpha = 10^{-4}$ to the three recipe partitions, at seeds 3 to~5, and uses the runs of the ablation as its arms without the loss. The \emph{large-model} experiment trains the three recipe partitions on the large model with 3B tokens and seeds 0 to~2. Only this experiment, its tuning, and a probe that repeats its first 3,000 steps use the large model.

Every comparison of an experiment is paired by seed. We report the mean of the three differences between runs with the same seed, followed by their range, which with three seeds equals the 95\% percentile bootstrap interval~\citep{efron1993introduction} and understates the uncertainty~\citep{agarwal2021precipice}. We call the paired difference of an arm from AdamW in a metric its \emph{excess} over AdamW\@. Before training each experiment, we listed the comparisons to test and, where we could, the sign we expected. We call these comparisons \emph{planned} and every other result exploratory, including the analyses chosen after seeing the runs and the comparisons that the plan wrote down but set apart from those to test. Two arms that do not differ give a difference with the same sign at all three seeds with probability $1/4$, and with a predicted sign with probability $1/8$. We apply no correction for the number of planned comparisons (\cref{app:program} counts how many held). After the main comparison had shown that Muon raised the span of the first MoE layer, this comparison was planned in the ablation, the bias-speed experiment, and the large-model experiment, and it held at all nine seeds of these three experiments, which two arms that do not differ would give with probability $2^{-9} \approx 0.002$. We report the values at equal tokens. Outside the stress test, every planned comparison except those of the loss and of the specialization, which is measured only at the last checkpoint, gave the same conclusion when read at equal validation loss. Every difference is computed at full precision and then rounded, so it can differ in its last digit from the difference of the rounded values.

Two exploratory probes repeat the first 3,000 steps of some runs with more metrics recorded, either unchanged or with one change. The \emph{probe of the ablation} repeats the runs of AdamW, Muon, and Muon at half rate unchanged at seed~3. At seeds 3 to~5 it \emph{centers} the input of the first router under Muon, subtracting its batch mean as \citet{mu2018allbutthetop} subtract the mean of word vectors, trains Muon with the fast bias, and trains the router on AdamW with the backbone on Muon at half the tuned rate of Muon. The \emph{probe of the large model} centers the input of the first router under both Muon partitions at seeds 0 to~2. At seed~0 it trains Muon with the fast bias and repeats the three recipe partitions unchanged. An unchanged probe run reproduces its full run bit for bit, so a changed run differs from its full run by the change alone. All runs were trained on NVIDIA RTX~5090 GPUs with deterministic algorithms, and none diverged.

\section{Results}
\label{sec:results}

\subsection{Muon Against AdamW}
\label{sec:rq1}

\Cref{tab:rq1} compares AdamW with Muon on the small model. Under Muon the update--load correlation fell from 0.38 to~0.02, so, as we had expected, experts with more tokens no longer received larger updates. Yet the span rose at every seed, from 0.105 to~0.137 averaged over the layers, and so did the integrated and the early load churn, against our predictions for all three (\cref{app:program}). Muon reached a lower validation loss, by 0.078, and a higher specialization. In every MoE layer but the first, the integrated overload stayed below 1.32 under both optimizers, so the bias kept the loads close to even and the optimizers differed in how far it had to move.

\begin{table}[ht]
  \centering
  \caption{The main comparison, on the small model with 1B tokens and seeds 0 to~2. Every metric is the integrated value averaged over the MoE layers unless the row names another layer or window. The validation loss is that of the last step and the specialization is measured at the last checkpoint. Each difference is Muon minus AdamW, followed by the range of the three paired differences.}
  \label{tab:rq1}
  \small
  \begin{tabular}{@{}lrrl@{}}
    \toprule
    Metric & AdamW & Muon & Difference \\
    \midrule
    Validation loss & 3.255 & 3.177 & $-0.078$ \seedrange{-0.086}{-0.062} \\
    Span, all layers & 0.105 & 0.137 & $+0.033$ \seedrange{+0.030}{+0.036} \\
    Span, first MoE layer & 0.084 & 0.194 & $+0.110$ \seedrange{+0.085}{+0.151} \\
    Update--load correlation & 0.379 & 0.024 & $-0.355$ \seedrange{-0.362}{-0.345} \\
    Update--load correlation, spectral norm & 0.102 & 0.035 & $-0.068$ \seedrange{-0.073}{-0.061} \\
    Load churn & 0.032 & 0.038 & $+0.006$ \seedrange{+0.004}{+0.008} \\
    Load churn, early & 0.049 & 0.064 & $+0.015$ \seedrange{+0.011}{+0.017} \\
    Warmup churn & 0.102 & 0.083 & $-0.019$ \seedrange{-0.020}{-0.018} \\
    Specialization & 0.105 & 0.118 & $+0.014$ \seedrange{+0.006}{+0.017} \\
    \bottomrule
  \end{tabular}
\end{table}

The difference in span was largest in the first MoE layer (\cref{fig:main}), where the span was 0.194 under Muon against 0.084 under AdamW, while the other six layers differed by 0.02 on average. In that layer the difference was largest early in training. Over steps 501 to~1,000 this span was 0.74 under Muon and 0.19 under AdamW, and over steps 8,001 to~15,259, about the second half of the run, it was still 0.139 against 0.062. We call the excess span of the first MoE layer with the backbone on Muon the \emph{rise}. In the tuning runs of the small model, the span averaged over the layers grew with the learning rate of Muon and stayed at 0.113 to~0.132 under AdamW (\cref{fig:main}, right), and the span of the first MoE layer grew from 0.158 to~0.959 under Muon and stayed at 0.080 to~0.144 under AdamW\@. In the tuning of the large model, this span grew from 0.087 to~0.987 under Muon and stayed at 0.160 to~0.203 under AdamW, so at a quarter of the selected rate of Muon it was below that of AdamW at every rate. Part of the rise may therefore come from the tuned rate of Muon, twice that of AdamW~(\cref{sec:rate}).

\subsection{The Router and the Backbone}
\label{sec:backbone}

The ablation crosses the optimizer of the router with that of the backbone (\cref{tab:rq3}, and \cref{fig:arms} in \cref{app:ablation}). The span of the first MoE layer followed the backbone. With the backbone on Muon, it was 0.198 with the router on Muon and 0.175 with the router on AdamW, and with the backbone on AdamW it was 0.099 and 0.095. Both Muon partitions exceeded both arms with the backbone on AdamW at every seed. Changing only the optimizer of the router moved this span by 0.023 or less on average, in a direction that varied with the seed, although the router on Muon raised the span averaged over all MoE layers at every seed with either backbone. The input cosine of the first router followed the backbone in the same way, about 0.4 with the backbone on Muon against 0.07 on AdamW\@. At these seeds, Muon again reached a lower loss than AdamW, by 0.077, and a higher span of the first MoE layer, by 0.103 (\cref{tab:rise}).

\begin{table}[t]
  \centering
  \caption{The arms of the ablation with the bias, on the small model with 1B tokens, as means over seeds 3 to~5. The span, the input cosine, the score offset, the early overload (steps 501 to~1,262), and the common share are measured in the first MoE layer, and the last two columns over all MoE layers. The loss is the validation loss at the last step, and the other metrics are integrated values unless the column names a window. The arm ``Muon, half rate'' trains with Muon at half its tuned rate, which is the rate of AdamW\@.}
  \label{tab:rq3}
  \small
  \setlength{\tabcolsep}{0pt}
  \begin{tabular*}{\textwidth}{@{\extracolsep{\fill}}l*{4}{wc{1.2cm}}wc{1.5cm}wc{1.2cm}wc{1.5cm}wc{1.4cm}@{}}
    \toprule
    & & & Input & Score & Early & Common & Span, & Warmup \\
    Arm & Loss & Span & cosine & offset & overload & share & all layers & churn \\
    \midrule
    AdamW & 3.253 & 0.095 & 0.071 & 0.095 & 1.14 & 0.129 & 0.103 & 0.102 \\
    AdamW, router on Muon & 3.248 & 0.099 & 0.074 & 0.100 & 1.16 & 0.151 & 0.117 & 0.092 \\
    Muon, router on AdamW & 3.177 & 0.175 & 0.391 & 0.137 & 6.63 & 0.497 & 0.123 & 0.122 \\
    Muon & 3.175 & 0.198 & 0.408 & 0.140 & 4.06 & 0.160 & 0.140 & 0.084 \\
    \midrule
    Muon, half rate & 3.166 & 0.119 & 0.401 & 0.100 & 1.73 & 0.161 & 0.115 & 0.074 \\
    \midrule
    Experts on Muon & 3.205 & 0.409 & 0.146 & 0.298 & 3.24 & 0.201 & 0.175 & 0.103 \\
    Attention on Muon & 3.220 & 0.130 & 0.053 & 0.109 & 2.06 & 0.213 & 0.118 & 0.104 \\
    Dense FFN on Muon & 3.227 & 0.094 & 0.033 & 0.090 & 1.24 & 0.169 & 0.110 & 0.100 \\
    Attention, dense FFN on Muon & 3.210 & 0.229 & 0.054 & 0.183 & 5.07 & 0.226 & 0.140 & 0.100 \\
    \bottomrule
  \end{tabular*}
\end{table}

\begin{table}[t]
  \centering
  \caption{The excess over AdamW across experiments. Each entry is the paired difference of an arm from AdamW with the same balancer, followed by its range over the seeds. The span, the early overload, and the input cosine are measured in the first MoE layer, the span and the input cosine as integrated values and the early overload over steps 501 to~1,262 (501 to~2,788 on the large model). The loss is the validation loss at the last step. The main comparison did not record the input cosine. The half rate is half the tuned rate of Muon, and the fast bias moves by $4u$ per step against $u$ for the usual bias.}
  \label{tab:rise}
  \footnotesize
  \setlength{\tabcolsep}{2pt}
  \begin{tabular*}{\textwidth}{@{\extracolsep{\fill}}>{\raggedright\arraybackslash\hangindent=0.8em\hangafter=1}p{2.32cm}llll@{}}
    \toprule
    Arm $-$ AdamW & Span & Early overload & Input cosine & Loss \\
    \midrule
    \multicolumn{5}{@{}l}{\emph{Main comparison, small model, seeds 0 to~2}} \\
    Muon & $+0.110$ \seedrange{+0.085}{+0.151} & $+3.15$ \seedrange{+1.47}{+4.58} & -- & $-0.078$ \seedrange{-0.086}{-0.062} \\
    \addlinespace[3pt]\multicolumn{5}{@{}l}{\emph{Ablation, small model, seeds 3 to~5}} \\
    Muon & $+0.103$ \seedrange{+0.045}{+0.174} & $+2.93$ \seedrange{+2.17}{+3.52} & $+0.338$ \seedrange{+0.324}{+0.345} & $-0.077$ \seedrange{-0.083}{-0.067} \\
    Muon, router on~AdamW & $+0.080$ \seedrange{+0.065}{+0.091} & $+5.50$ \seedrange{+5.06}{+5.75} & $+0.321$ \seedrange{+0.313}{+0.335} & $-0.075$ \seedrange{-0.081}{-0.066} \\
    Muon, half rate & $+0.024$ \seedrange{+0.010}{+0.037} & $+0.59$ \seedrange{+0.48}{+0.78} & $+0.330$ \seedrange{+0.321}{+0.342} & $-0.087$ \seedrange{-0.095}{-0.075} \\
    \addlinespace[3pt]\multicolumn{5}{@{}l}{\emph{Bias speed, small model, seeds 6 to~8}} \\
    Muon, usual bias & $+0.075$ \seedrange{+0.041}{+0.095} & $+2.84$ \seedrange{+2.20}{+3.34} & $+0.329$ \seedrange{+0.324}{+0.332} & $-0.082$ \seedrange{-0.094}{-0.069} \\
    Muon, fast bias & $+0.040$ \seedrange{+0.013}{+0.066} & $+0.54$ \seedrange{+0.46}{+0.69} & $+0.324$ \seedrange{+0.320}{+0.327} & $-0.082$ \seedrange{-0.088}{-0.071} \\
    \addlinespace[3pt]\multicolumn{5}{@{}l}{\emph{Balance loss, small model, seeds 3 to~5}} \\
    Muon & $+0.081$ \seedrange{+0.045}{+0.121} & $+1.67$ \seedrange{+1.41}{+2.00} & $+0.338$ \seedrange{+0.329}{+0.347} & $-0.077$ \seedrange{-0.080}{-0.073} \\
    Muon, router on~AdamW & $+0.055$ \seedrange{+0.039}{+0.073} & $+5.11$ \seedrange{+4.97}{+5.34} & $+0.328$ \seedrange{+0.321}{+0.338} & $-0.076$ \seedrange{-0.079}{-0.071} \\
    \addlinespace[3pt]\multicolumn{5}{@{}l}{\emph{Large model, seeds 0 to~2}} \\
    Muon & $+0.152$ \seedrange{+0.087}{+0.240} & $+5.19$ \seedrange{+4.46}{+6.10} & $+0.525$ \seedrange{+0.495}{+0.583} & $-0.045$ \seedrange{-0.048}{-0.040} \\
    Muon, router on~AdamW & $+0.125$ \seedrange{+0.068}{+0.186} & $+6.09$ \seedrange{+5.32}{+6.57} & $+0.642$ \seedrange{+0.604}{+0.714} & $-0.041$ \seedrange{-0.044}{-0.037} \\
    \bottomrule
  \end{tabular*}
\end{table}

\subsection{The Shared Direction}
\label{sec:direction}

Across the 27 runs of the ablation with the bias, the logarithm of the score offset of the first MoE layer explained 93\% of the variance of the logarithm of its span, and the logarithm of the input cosine 11\% (\cref{fig:geometry} in \cref{app:ablation}). In every run of the probe of the ablation that did not center the input, the correlation across experts between the bias and the offset, averaged over steps 501 to~3,000, was $-0.85$ to~$-0.70$. Since the bias exists to cancel the preference that the offsets create, these two results mainly show that the span measures the score offset. They leave open what sets the score offset. By \cref{eq:offset}, the spread of the offsets is the product of two factors, the length of the mean input relative to the average length of a single input and the spread of the rows of the router along the mean input (\cref{app:ablation} defines both). In the probe of the ablation, over steps 501 to~3,000 and with the backbone on Muon at its tuned rate or at half that rate, the length of the mean input of the first router was 0.63 to~0.67 times the average length of a single input, against 0.21 in the run of AdamW at seed~3. Relative to that run, the spread of the rows along the mean input was 3.0 at the tuned rate of Muon and 1.4 at half the rate. The length of the mean input followed the backbone, while the spread of the rows changed with the rate of Muon.

\begin{figure}[t]
  \centering
  \includegraphics{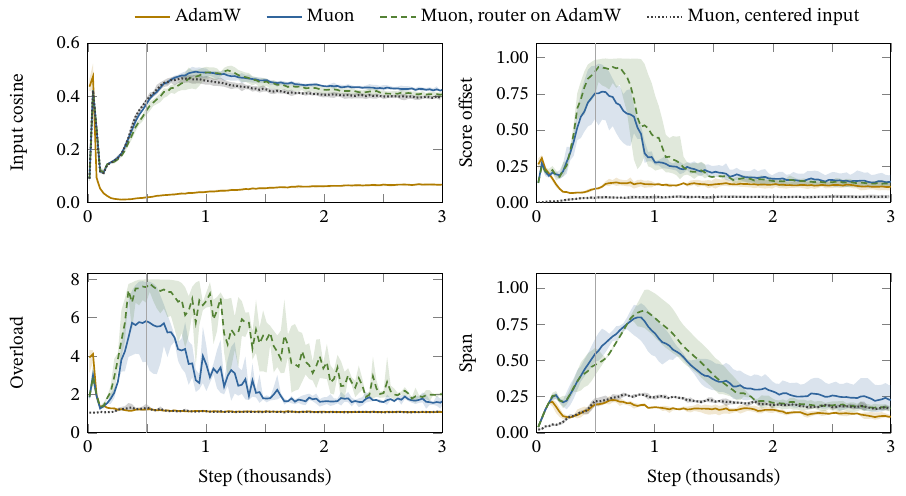}
  \caption{The input cosine, the score offset, the overload, and the span of the first MoE layer over the first 3,000 steps of the ablation on the small model, in bins of 30 steps, under AdamW, Muon, and Muon with the router on AdamW, and in the probe that repeats the runs of Muon with the batch mean subtracted from the input of the first router. The input cosine of the probe is measured before the subtraction. Lines are the means over seeds 3 to~5 and bands their range. The thin gray vertical line marks the end of the warmup.}
  \label{fig:early}
\end{figure}

\Cref{fig:early} follows the first 3,000 steps. After a brief peak near step~45, the input cosine was 0.2 or less at step~100 on every line of the figure. It stayed low under AdamW, while with the backbone on Muon it rose again, crossed 0.2 at steps 310 to~340 and averaged 0.43 to~0.48 over steps 501 to~1,262. Without centering, the score offset rose with it and peaked at 0.64 to~0.99, at steps 490 to~720. The bias lagged behind, since a difference between two biases grows by at most $2u = 0.002$ per step and a span of 0.8 takes at least 400 steps to build. The span peaked 200 to~290 steps after the score offset, at 0.74 to~0.99, and until then the favored experts received more than their fair share. The early overload was 4.06 under Muon and 6.63 with the router on AdamW, against 1.14 under AdamW\@. Later the score offset fell while the input cosine stayed near 0.4, and over the last 1,000 steps of the full runs the rise was about a tenth of its value over steps 501 to~1,262.

To test whether the shared direction causes the overload, the probe of the ablation repeated the first 3,000 steps of the three runs of Muon with the input of the first router centered. The router then no longer sees the shared direction, while the experts of every layer receive their inputs unchanged. With centering, the early overload was 1.11 to~1.16, about that of AdamW, against 3.30 to~4.66 without it, and over steps 501 to~3,000 the rise fell by 79\% (\cref{fig:early}). The input cosine before centering stayed near 0.42, and the loss at step~3,000 changed by $-0.0012$ to~$+0.0009$.

No partition with only part of the backbone on Muon reproduced the shared direction (\cref{tab:rq3}). The input cosine was 0.146 with the routed experts on Muon and at most 0.054 in the other three, below that of both Muon partitions at every seed. Two of these partitions still raised the span more than the whole backbone did at every seed, through a large score offset, to~0.409 with the routed experts on Muon and 0.229 with the attention and the dense FFN (\cref{app:ablation}).

\subsection{The Learning Rate of Muon}
\label{sec:rate}

At half its tuned rate, which is the rate of AdamW, Muon still raised the span of the first MoE layer at every seed, but by 0.024 against 0.103 at the tuned rate, or 25\% of the span of AdamW against 108\% (\cref{tab:rise}). Averaged over the layers, its excess over AdamW fell from 0.038 to~0.012 and was positive at two of the three seeds. The half rate reached a lower loss than the tuned rate at every seed, by 0.008 to~0.012, although the tuning with 0.5B tokens had favored the tuned rate by 0.003. This agrees with the fall of the best rate as runs grow longer~\citep{bjorck2025horizons}. The half rate kept the shared direction, with an input cosine of 0.40, and lowered the spread of the rows of the router along it (\cref{sec:direction}). It brought the score offset down to about that of AdamW and the early overload from 4.06 to~1.73 (\cref{tab:rq3}). The large model, whose rates were scaled down for its longer runs (\cref{sec:rates}), still showed the rise at every seed (\cref{sec:robustness}).

\subsection{Other Balancers and the Large Model}
\label{sec:robustness}

The fast bias halved the rise, from 0.075 to~0.040 (\cref{tab:rise}), and cut the excess early overload from 2.84 to~0.54, but the span stayed higher under Muon at every seed. Under either optimizer, the fast bias changed the input cosine and the loss by less than 0.01 on average. In the probe of the ablation, the fast bias left the length of the mean input almost unchanged, and the spread of the rows of the router along it was 0.69 to~1.57 times that of the run of AdamW at seed~3 (\cref{app:ablation}).

With the balance loss added to the bias, the span stayed higher than under AdamW at every seed, by 0.081 under Muon and 0.055 with the router on AdamW, and the input cosine by 0.34 and 0.33. For a softmax router trained only to balance its loads, with small logits, \citet[Prop.~2]{wang2026myth} show that a simplified balance loss, the squared distance of the batch-mean softmax output from uniform, makes a shared direction add the same logit to every expert, and they expect this only in part when the loss is auxiliary. In our terms, a balance loss should lower the score offset. Our balance loss lowered the score offset of all three recipe partitions and the early overload of both Muon partitions at every seed, as this predicts. Its effects on the input cosine and the common share varied in sign over the seeds. We had predicted that it would also lower the span under Muon, and it did at two of the three seeds (\cref{app:program}). With the router on AdamW, the balance loss lowered this span at every seed, by 0.037 to~0.054.

Without the bias, Muon kept the usage of the experts more even than AdamW on average over the layers, as \citet{wang2026depth} found, but less even in the first MoE layer, where the early overload under Muon was 7.7 to~7.9 (\cref{app:ablation}).

On the large model, with 3B tokens and its own tuned rates, Muon raised the span of the first MoE layer at every seed, by 0.152 on average, or 112\% of the span of AdamW against 108\% in the ablation on the small model. The input cosine was 0.61 under Muon and 0.73 with the router on AdamW, against 0.41 and 0.39 on the small model, and the early overload was 6.3 and 7.2, near the ceiling of 8. Nine of the ten planned comparisons of this experiment (\cref{app:program}) held at every seed, and \cref{sec:router} gives the exception. The first MoE layer of the large model was still overloaded at the end of training, with its busiest expert at 2.2 to~2.8 times its fair share over the last 10,000 steps under both Muon partitions against 1.07 under AdamW\@. In the unchanged runs of the probe of the large model at seed~0, the scores of this layer sank toward zero, into the flat part of the sigmoid, under both Muon partitions, and over steps 2,001 to~3,000 the bias made most of the choices of experts. At seeds 0 to~2, centering the input of the first router kept the scores off the flat part and removed the early overload (\cref{app:large}).

\subsection{The Optimizer of the Router}
\label{sec:router}

With the backbone on Muon, moving the router to AdamW made the loads shift more during the warmup and raised the loss slightly (\cref{tab:router}). On both model sizes and with the balance loss, the warmup churn and the loss were higher at every seed, the loss by 0.001 to~0.004. The comparison of the warmup churn was planned in the ablation and the large-model experiment, and that of the loss in the ablation. The early overload was higher at every seed on the small model, and it became a planned comparison of the large-model experiment. On the large model, with both partitions at 5.6 to~7.7 times the fair share, near the ceiling of 8, it was higher at two seeds and lower at one, which makes it the one planned comparison of that experiment that did not hold at every seed. In a window we chose after seeing this result, steps 301 to~500, before either partition approached the ceiling, the overload on the large model was higher with the router on AdamW at every seed. The large-model experiment had 2, 6, and 2 loss spikes in the runs of Muon with the router on AdamW at the three seeds and none in the runs of AdamW or Muon (\cref{app:large}, with spikes defined in \cref{app:stress}). At nine of these ten spikes, the span of the first MoE layer was about $2u$ or less below its ceiling. In the tuning of the large model, which had no run with the router on AdamW, only Muon at twice the selected rate had loss spikes, again with this span about $2u$ or less below its ceiling (\cref{app:large}).

\begin{table}[t]
  \centering
  \caption{The router on AdamW minus the router on Muon, both with the backbone on Muon, as paired differences with their range over the seeds, which are 3 to~5 on the small model with 1B tokens and 0 to~2 on the large model with 3B tokens. The early overload, the common share, and the span are measured in the first MoE layer, the early overload over steps 501 to~1,262 (501 to~2,788 on the large model) and the other two as integrated values. The warmup churn is averaged over all MoE layers, and the loss is the validation loss at the last step.}
  \label{tab:router}
  \small
  \begin{tabular}{@{}llll@{}}
    \toprule
    & & Small model, & \\
    Metric & Small model & balance loss & Large model \\
    \midrule
    Warmup churn & $+0.038$ \seedrange{+0.038}{+0.039} & $+0.045$ \seedrange{+0.043}{+0.046} & $+0.038$ \seedrange{+0.037}{+0.039} \\
    Early overload & $+2.57$ \seedrange{+1.96}{+3.52} & $+3.44$ \seedrange{+2.96}{+3.93} & $+0.91$ \seedrange{-0.78}{+1.94} \\
    Common share & $+0.337$ \seedrange{+0.328}{+0.342} & $+0.339$ \seedrange{+0.334}{+0.346} & $+0.377$ \seedrange{+0.339}{+0.429} \\
    Span & $-0.023$ \seedrange{-0.092}{+0.020} & $-0.025$ \seedrange{-0.048}{-0.005} & $-0.027$ \seedrange{-0.119}{+0.058} \\
    Loss & $+0.002$ \seedrange{+0.001}{+0.004} & $+0.001$ \seedrange{+0.001}{+0.002} & $+0.004$ \seedrange{+0.003}{+0.004} \\
    \bottomrule
  \end{tabular}
\end{table}

In the stress test, at every multiple of the tuned rates from 1 to~16, the warmup churn was highest with the router on AdamW and lowest under Muon, with ranges over the seeds that did not overlap, while the mean losses of the two Muon partitions differed by less than 0.0022 and no run diverged (\cref{app:stress}). From the report of Qwen3.8-Flash-Next, we had expected the router on AdamW to lower the warmup churn and the Muon partitions to be less sensitive to the learning rate than AdamW, and both expectations failed (\cref{app:program}). The stress test came first, so the ablation and the large-model experiment predicted this higher warmup churn.

The common share suggests why the router on AdamW shifted the loads more. When the inputs of the router share a direction, the gradient of every row has a large component along it, and a step along this direction changes the logit of an expert by about the same amount for every token. AdamW normalizes its update entry by entry and keeps much of this component, while Muon makes the singular values of its update nearly equal, and on Muon the first router had a common share of 0.12 to~0.24 in our runs. With the backbone on Muon, the common share of the first router was 0.497 on AdamW against 0.160 on Muon (\cref{tab:rq3}). The higher overload with the router on AdamW remained when the router and the backbone trained at the same rate. In the probe of the ablation, the router on AdamW with the backbone on Muon at half the tuned rate of Muon, which is the rate of AdamW, gave an early overload of 4.2 to~5.5, against 1.6 to~1.9 in the three runs of the ablation with Muon at half rate. The overload does not seem to account for the higher loss. In the probe of the large model, centering removed the overload under both Muon partitions, and the loss with the router on AdamW stayed 0.005 to~0.007 above that of Muon at step~3,000.

\section{Discussion and Limitations}
\label{sec:discussion}

\subsection{An Account of the Rise}
\label{sec:mechanism}

Our results support the following account. With the backbone on Muon, the inputs of the first router share a direction, and by \cref{eq:offset} their mean gives every token the same preference among the experts, in proportion to how much the rows of the router differ along it. Each bias moves by $u$ per step, so the favored experts are overloaded until the biases catch up. The span followed the score offset across the arms, and in the planned comparisons that crossed the optimizers of the router and the backbone, the span and the shared direction both followed the backbone. In the 3,000-step probes, centering the input of the first router removed the early overload on both model sizes and changed the loss by less than 0.003.

Three questions remain open. First, we do not know what spreads the rows of the router along the shared direction. On the small model, half the tuned rate of Muon and the fast bias each lowered this spread and left the length of the mean input almost unchanged, and in both cases the early overload stayed below twice the fair share. This suggests that the rows move apart because the loads are uneven, since a row receives gradient only from the tokens routed to its expert, but we did not test this. Second, we do not know how the routed experts alone on Muon gave the largest span of the ablation at an input cosine of 0.146. The control, which gives every expert an update of the same size without the orthogonalization, had a higher span than AdamW at every rate above its tuned rate (\cref{app:tuning}), so updates of equal size may matter, and training only the routed experts with the control would test this. Third, and most important, we do not know why, of the partitions we trained, only the whole backbone on Muon raised the input cosine of the first router to about 0.4. With only the attention and the dense FFN on Muon, every weight matrix that computes the input of this router except the embedding table was on Muon, yet the input cosine stayed at 0.054 while the early overload reached 5.07. The shared direction therefore seems to need the routed or the shared experts on Muon, although the experts shape this input only through the gradients they send back. The routed experts alone were not enough, and no partition other than the whole backbone put the shared experts on Muon.

\subsection{The Router and the First MoE Layer}
\label{sec:where}

At our scale, the comparisons favor the router on Muon when the backbone is on Muon, as do the runs of \citet{lau2026symmetry} without a balancing bias, while Qwen3.8-Flash-Next keeps its router on AdamW because, according to its report, Muon made it unstable early in training. Qwen3.8-Flash-Next is far larger, with 125B parameters of which 6B are active, and its own stress test used a model with 25B parameters. Its report shows no data on the router and states neither the balancer nor the loads, so we cannot tell whether the early instability it describes is the warmup churn we measure. The released configuration~\citep{qwen2026flashnextcard} lists a balance loss and no bias, so its router may face a different balancer than ours. The 60B-parameter models of \citet{jin2026silent} are balanced by the bias, as ours are. With the backbone on Muon, their router on AdamW left the first MoE layer less balanced than their router on Muon, early in training and still at its end, and reached a final validation loss about 0.001 lower, in one run each. In their seventh MoE layer, the router on AdamW was again less balanced early in training but more balanced at the end. With the backbone on AdamW, moving the router to Muon, which no recipe we cite does, changed the span of the first MoE layer by $+0.004$ and the loss by $-0.005$ on average, with signs that varied over the seeds. In the stress test, the largest absolute router logit under Muon at 16 times the tuned rates averaged 98 over the seeds and reached 170 at one seed, while that of a router on AdamW stayed below 36 (\cref{app:stress}). Our sigmoid router has no softmax and runs in single precision, and the stress test had too few loss spikes to tell whether such logits harm training, but in a softmax router computed in bfloat16 they could cause the roundoff errors that the router z-loss reduces~\citep{zoph2022stmoe}.

In our models, which have no capacity limit, the overload of the first MoE layer had no cost that we could measure. The only possible exception is the loss spikes with the backbone on Muon. At the tuned rates they came only on the large model with the router on AdamW, and otherwise only at 2 to~16 times the selected rates, in the tunings and the stress test. At every one of them except a spike with the router on AdamW in the stress test, the span of that layer was 0.99 or more, near its ceiling (\cref{app:stress,app:large}). When experts sit on separate devices, busy experts slow down training~\citep{lepikhin2021gshard}, and with a capacity limit the tokens beyond it are dropped~\citep{fedus2022switch}, so four to seven times the fair share for hundreds of steps would cost time or tokens. Our models, like DeepSeekMoE and IQuest-Q1~\citep{dai2024deepseekmoe, yang2026iquest}, have a dense block before the first MoE layer, and under Muon the MoE layer after it was still the hardest to balance. Centering removed the overload, but we tested it only over 3,000 steps and in one layer, and the batch mean makes the routing of a token depend on its batch, so a practical version would subtract a running mean. The fast bias lowered the overload, but in the probe of the large model, at one seed, the loads of the first MoE layer still shifted about three times as much as under AdamW (\cref{app:large}).

\subsection{Limitations}
\label{sec:limitations}

Our models have 55M and 108M active parameters, about 50 to~2,000 times fewer than the models whose recipes motivate our question, and they train on at most 3B tokens, about four orders of magnitude fewer. At about 60B parameters, 1B of them active, the first MoE layer was less balanced early in training under Muon than under AdamW and least balanced with the router on AdamW~\citep{jin2026silent}, the same order as the early overload in our ablation (\cref{tab:rq3}), but that comparison has one run per arm and measures neither the bias nor the inputs of the router. In trained Moonlight and DeepSeek-V2-Lite, which have almost the same architecture but differ in optimizer, balancer, router scoring function, tokenizer, and training data, the mean pairwise cosine similarity of the hidden states is higher in Moonlight, trained with Muon, in the lowest 20 of the 26 layers plotted and lower in the last three~\citep[Fig.~15]{wang2026myth}, so at scale a stronger shared direction under Muon may persist mainly in the lower layers. In an audit of released models~\citep[App.~F]{jin2026silent}, adding the bias to the scores left the expert with the highest score out of the selection for most tokens in some layers of Kimi~K2~\citep{kimi2025k2}, Kimi~K3, and GLM-5, all trained with Muon, and for almost no tokens in any layer of DeepSeek-V3, trained with AdamW, or of DeepSeek-V4-Pro, trained with Muon. In Kimi~K2 and GLM-5 this held in the first MoE layer, but the models differ in more than the optimizer. The input cosine and the early overload of the first MoE layer were higher on the large model, which trained on three times as many tokens as the small one, and we cannot say whether the overload of this layer would settle in a much longer run. In Nemotron~3 Ultra, trained with AdamW, the busiest expert of the first MoE layer reached about 12 times its fair share as its training grew unstable, so in large models this layer may be hard to balance under either optimizer, at least in unstable runs~(\cref{sec:related}).

We train one architecture on one dataset, with 32 routed experts of which a sigmoid router selects~4, a shared expert, one dense block before the first MoE layer, and no capacity limit. The experts of the large model are narrower than its width, 512 against 768, while those of the small model are wider, 768 against 512, so the two sizes differ in more than scale. The sigmoid bounds the scores and gives the span its ceiling, which the span of the first MoE layer nearly reached at some steps on both models, whereas the scores of DeepSeek-V4 have no upper bound~\citep{deepseek2026v4}. We test one balancer, the bias, at two update rates, and add a balance loss at one weight. We use one batch size, 65,536 tokens, which sets the noise in the loads that the bias reacts to, and one warmup of 500 steps, inside which the shared direction appeared. We evaluate the models only by their validation loss, which in small models trained on few tokens may not tell routing methods apart~\citep{falke2026testbed}.

The rates of the small model were tuned with one seed and 0.5B tokens and used with 1B, where half the tuned rate of Muon trained better and removed most of the rise. We did not retune AdamW on the small model at 1B tokens or train the large model with 3B tokens at a lower rate of Muon. \citet{bjorck2025horizons} find a steeper fall of the best rate with the number of tokens in their models of 50M and 125M parameters than in larger ones, and ours have 55M and 108M active parameters, so the rates of the large model may be above the best for both optimizers. Each comparison of an experiment rests on three seeds, with no correction for the number of comparisons, and several results of the probes rest on one seed. The probes cover only the first 3,000 steps and, like most of the other exploratory analyses, were chosen after seeing the runs.

\section{Conclusion}
\label{sec:conclusion}

In MoE language models with 55M and 108M active parameters, balanced by the auxiliary-loss-free bias, Muon raised the span of the first MoE layer to about twice that of AdamW on average, at learning rates tuned on shorter runs. This span rose with Muon on the backbone, whatever optimizer trained the router, and the backbone on Muon gave the inputs of the first router a shared direction. On the small model, most of the rise disappeared at half the tuned rate of Muon, which trained better at 1B tokens. Centering the input of the first router removed the early overload in 3,000-step probes and changed the loss by less than 0.003. With the backbone on Muon, moving the router to AdamW shifted the loads more during the warmup and, at the tuned rates, raised the loss slightly. The next questions are whether these effects grow or settle at production scale and whether a running mean can replace the batch mean in centering.

{\renewcommand{\sectionpage}{}\section*{Acknowledgments}}

I thank Daniel Hernández-Lobato for supervising the thesis and for his comments on its writing.

\bibliography{references}

\appendix

\section{Configuration}
\label{app:configuration}

\Cref{tab:architecture} gives the sizes of the two models. The output head is not tied to the embedding table. Every normalization layer is an RMSNorm with $10^{-5}$ added to the mean square and gains initialized to one. One precedes the attention and one the FFN or MoE layer of each block, and one precedes the output head. The attention normalizes the queries and the keys with an RMSNorm whose gain vector is shared by the heads, and its gate multiplies the attention output entry by entry by the sigmoid of another projection of the hidden state, before the output projection. The rotary embeddings have base 10,000. No linear map has a bias term, and every weight matrix, including the embedding table and the output head, is initialized from a normal distribution with standard deviation 0.02. The router computes the scores, the selection, and the mixing weights of \cref{eq:moe} in single precision.

\begin{table}[ht]
  \centering
  \caption{The two models. An expert, routed or shared, is a SwiGLU FFN with three matrices. The active parameters are those one token uses, which in every MoE layer are the router, the shared expert, and $k$ routed experts, and they exclude the embedding table and the output head.}
  \label{tab:architecture}
  \small
  \begin{tabular*}{\textwidth}{@{\extracolsep{\fill}}lrr@{}}
    \toprule
    & Small model & Large model \\
    \midrule
    Blocks (dense, MoE) & 8 (1, 7) & 12 (1, 11) \\
    Width $d$, attention heads, head width & 512, 8, 64 & 768, 12, 64 \\
    Hidden width of the dense FFN & 2,048 & 3,072 \\
    Routed experts $E$, selected $k$, shared experts & 32, 4, 1 & 32, 4, 1 \\
    Hidden width of an expert & 768 & 512 \\
    Vocabulary & 16,384 & 16,384 \\
    Context & 1,024 & 1,024 \\
    Batch (sequences) & 64 & 64 \\
    Active parameters, total parameters & 55M, 303M & 108M, 496M \\
    Tokens (steps) per run & 0.5B (7,629) or 1B (15,259) & 1B (15,259) or 3B (45,776) \\
    \bottomrule
  \end{tabular*}
\end{table}

\Cref{tab:optimizer} lists the settings of the optimizers. Outside the stress test, the learning rate of step $t$ in a run of $S$ steps is
\begin{equation*}
  \eta_t = \frac{t}{500}\, \eta \quad (t \leq 500), \qquad \eta_t = \eta \biggl( 0.1 + 0.45 \biggl( 1 + \cos \frac{\pi (t - 500)}{S - 500} \biggr) \biggr) \quad (t > 500),
\end{equation*}
where $\eta$ is the peak rate of the optimizer, so the ratio between the rates of AdamW and Muon is the same at every step. Our Muon keeps an exponential moving average of the gradients in place of the momentum $\mathbf{M}_t$ of \cref{eq:muon}, as PyTorch does. The average is a positive multiple of the momentum, and neither the orthogonalization nor the rescaling of the control changes when its argument is multiplied by a positive number, so the updates are those of \cref{eq:muon}. The Newton--Schulz iteration divides the matrix by its Frobenius norm and applies
\begin{equation*}
  \mathbf{X} \leftarrow c_1 \mathbf{X} + c_3 (\mathbf{X} \mathbf{X}^{\top}) \mathbf{X} + c_5 (\mathbf{X} \mathbf{X}^{\top})^2 \mathbf{X}, \qquad (c_1, c_3, c_5) = (3.4445, -4.7750, 2.0315),
\end{equation*}
five times, with the coefficients of \citet{jordan2024muon}. In exact arithmetic, these five steps bring every singular value that starts above about 0.002 into roughly 0.7 to~1.2. The iteration runs in bfloat16 on all the matrices of one shape at once.

\begin{table}[ht]
  \centering
  \caption{The settings of the optimizers, the learning rates, and the balancers. The rates are peak rates.}
  \label{tab:optimizer}
  \small
  \begin{tabular}{@{}l>{\raggedright\arraybackslash}p{12.25cm}@{}}
    \toprule
    Setting & Value \\
    \midrule
    Muon & Nesterov momentum $\mu = 0.95$, five Newton--Schulz steps, factor~$0.2 \sqrt{\max(p, q)}$, $\lambda = 0.1$ \\ \addlinespace[2pt]
    AdamW & $\beta_1 = 0.9$, $\beta_2 = 0.95$, $\epsilon = 10^{-8}$, $\lambda = 0.1$ on every weight matrix, including the embedding table and the output head, and none on the gains \\ \addlinespace[2pt]
    Control & the Nesterov momentum of Muon rescaled to the Frobenius norm $\sqrt{\min(p, q)}$, then the factor and $\lambda$ of Muon \\ \addlinespace[2pt]
    Tuning grids & AdamW from $2.5 \times 10^{-4}$ to~$4 \times 10^{-3}$ (five rates), Muon from $2 \times 10^{-3}$ to~$3.2 \times 10^{-2}$ (five), the control from $2.5 \times 10^{-4}$ to~$8 \times 10^{-3}$ (six), each rate twice the one before \\ \addlinespace[2pt]
    Rates, small model & AdamW $2 \times 10^{-3}$, Muon $4 \times 10^{-3}$, the control $5 \times 10^{-4}$ (tuning only) \\ \addlinespace[2pt]
    Rates, large model & tuned with 1B tokens, AdamW $10^{-3}$ and Muon $2 \times 10^{-3}$, and used with 3B tokens, AdamW $7 \times 10^{-4}$ and Muon $1.4 \times 10^{-3}$ \\ \addlinespace[2pt]
    Clipping & 1.0 for each optimizer, and in the stress test twice the median gradient norm after the warmup of a calibration run of the same partition, from 0.170 to~0.277 \\ \addlinespace[2pt]
    Balancer & \cref{eq:bias} after every optimizer step with $u = 0.001$, or $0.004$ with the fast bias \\ \addlinespace[2pt]
    Balance loss & $\alpha = 10^{-4}$, the loss of DeepSeek-V3 computed in each sequence of 1,024 tokens on the scores normalized to sum to one, averaged over the sequences and summed over the MoE layers \\
    \bottomrule
  \end{tabular}
\end{table}

\section{Runs and Planned Comparisons}
\label{app:program}

\Cref{tab:runs} lists the runs of every experiment. Every one of the 145 runs finished its planned steps. They were trained on rented Vast.ai hosts with four NVIDIA RTX~5090 GPUs each, one run per GPU at a time, in 73 hours and 12 minutes of wall-clock time, and the training steps took 252.6 GPU-hours, without counting the setup of the hosts, the evaluation of the specialization on the four domains, and the exploratory runs. Training uses bfloat16 mixed precision with the master weights and the routers in single precision, \texttt{torch.compile}~\citep{ansel2024pytorch2} in PyTorch 2.14~\citep{paszke2019pytorch}, and deterministic algorithms, so a run trained again with the same seed on the same GPU model reproduces its numbers bit for bit. Computing the metrics costs about 14\% of the time of a run on the small model and 12\% on the large one, and two runs that differed only in which metrics they computed had bit-identical losses.

\begin{table}[ht]
  \centering
  \caption{The runs of every experiment. The tuning trains one run per rate and a run of AdamW with $\beta_2 = 0.98$, and the stress test adds a calibration run of each partition. The 145 runs above the rule do not include the 30 exploratory runs.}
  \label{tab:runs}
  \small
  \setlength{\tabcolsep}{4pt}
  \begin{tabular}{@{}lll>{\raggedright\arraybackslash}p{8.95cm}r@{}}
    \toprule
    Experiment & Model & Tokens & Arms and seeds & Runs \\
    \midrule
    Tuning & small & 0.5B & AdamW and Muon at five rates, the control at six, AdamW with $\beta_2 = 0.98$, seed~0 & 17 \\ \addlinespace[2pt]
    Main comparison & small & 1B & AdamW and Muon, seeds~0~to~2 & 6 \\ \addlinespace[2pt]
    Stress test & small & 0.5B & three recipe partitions at five multiples and Muon with the fast bias at multiple~4, seeds~0~to~2, and three calibration runs & 51 \\ \addlinespace[2pt]
    Ablation & small & 1B & eight partitions, Muon at half rate, and AdamW and Muon without the bias, seeds~3~to~5 & 33 \\ \addlinespace[2pt]
    Bias speed & small & 1B & AdamW and Muon with the usual and the fast bias, seeds~6~to~8 & 12 \\ \addlinespace[2pt]
    Balance loss & small & 1B & AdamW, Muon, and Muon with the router on AdamW, all with the balance loss, seeds~3~to~5 & 9 \\ \addlinespace[2pt]
    Tuning, large & large & 1B & AdamW and Muon at four rates each, seed~0 & 8 \\ \addlinespace[2pt]
    Large model & large & 3B & AdamW, Muon, and Muon with the router on AdamW, seeds~0~to~2 & 9 \\
    \midrule
    Exploratory & both & ${\leq}\,$0.5B & 8 pilot runs with seed~0 before the ablation, one per partition, three of them to about step~3,000, and the probes of the ablation (12 runs) and of the large model (10 runs), to step~3,000 & 30 \\
    \bottomrule
  \end{tabular}
\end{table}

In the main comparison, three of the seven planned comparisons held at every seed, a lower loss, a lower update--load correlation, and a higher specialization under Muon. Three failed with the opposite sign at every seed, the predictions that Muon would lower the span over all MoE layers, the integrated load churn, and the early load churn. The seventh, that Muon would lower the early dispersion of the update norms of the rows of the router, their standard deviation across experts divided by their mean, held on average but not at every seed. The later experiments measured this dispersion again outside the planned comparisons. It was lower under Muon than under AdamW at every seed in the bias-speed experiment. In the ablation and on the large model it was lower under Muon than with the router on AdamW at every seed. Against AdamW, Muon had about the same mean in the ablation, lower at two seeds and higher at the third, and was higher at every seed on the large model. The two planned comparisons of the stress test, whose expected signs came from the report of Qwen3.8-Flash-Next, predicted that the Muon partitions would have a lower learning-rate sensitivity than AdamW and that the router on AdamW would lower the warmup churn. Both failed, since the Muon partitions had a higher learning-rate sensitivity (\cref{app:stress}) and the router on AdamW raised the warmup churn at every multiple. The plan expected the two Muon partitions to have about the same learning-rate sensitivity, which held, at 0.241 and 0.238. All eleven planned comparisons of the ablation that had an expected sign and all nine of the bias-speed experiment held at every seed, as did nine of the ten of the large-model experiment and two of the three of the balance-loss experiment.

Of the eleven planned comparisons of the ablation with an expected sign, three predicted that both Muon partitions would have a higher span and a higher input cosine in the first MoE layer than both arms with the backbone on AdamW, and that each partition with part of the backbone on Muon would have a lower input cosine there than both Muon partitions. Two predicted a higher span of the first MoE layer than AdamW, one for Muon at half rate and one for both the partition with only the routed experts on Muon and the one with only the attention and the dense FFN on Muon. The other six predicted that Muon would have a higher span over all MoE layers, a lower loss, and a higher specialization than AdamW, that the router on AdamW would have a higher warmup churn than AdamW and Muon, that, among the arms with the bias at the tuned rates, the three with the routed experts on Muon would have a lower update--load correlation than the five with them on AdamW, and that without the bias Muon would have a higher final usage perplexity than AdamW\@. The nine planned comparisons of the bias-speed experiment with an expected sign predicted, as in the ablation, that with the usual bias Muon would have a higher span and a higher input cosine in the first MoE layer, a higher span over all MoE layers, a lower loss, a higher specialization, and a lower update--load correlation than AdamW, and that with the fast bias Muon would have a higher span of the first MoE layer than AdamW and a lower span of that layer and a lower early overload than with the usual bias. Of the three planned comparisons of the balance-loss experiment with an expected sign, the two that held predicted that, with the balance loss, both Muon partitions would have a higher span and a higher input cosine in the first MoE layer than AdamW\@. The third, that the balance loss would lower the span of that layer under Muon, failed at one seed by 0.0015. Of the ten planned comparisons of the large-model experiment, seven predicted that both Muon partitions would have a higher span of the first MoE layer, a higher input cosine, a higher early overload, and a lower update--load correlation than AdamW, and that Muon would have a lower loss, a higher span over all MoE layers, and a higher specialization than AdamW\@. The other three predicted that the router on AdamW would have a higher warmup churn and a higher common share than the other two partitions and a higher early overload than Muon. Only the last did not hold at every seed (\cref{sec:router}).

Six planned comparisons had no expected sign, apart from the other metrics of each arm of the balance-loss experiment with and without the balance loss, which the plan compared without an expected sign. In the ablation, the loss was higher with the router on AdamW than under Muon at every seed, by 0.0012 to~0.0036. In the bias-speed experiment, the fast bias lowered the span of the first MoE layer under AdamW at every seed, by 0.002 to~0.014. In the balance-loss experiment, the balance loss lowered the span of the first MoE layer under AdamW at every seed, by 0.011 to~0.031, and changed the validation loss of AdamW, Muon, and the router on AdamW by $+0.0003$, $+0.0008$, and $-0.0004$ on average, with signs that varied over the seeds.

\section{Additional Results}
\label{app:results}

\subsection{Tuning}
\label{app:tuning}

The final validation losses at the tuned rates were 3.386 for AdamW, 3.300 for Muon, and 3.396 for the control, and AdamW with $\beta_2 = 0.98$, a check of its second-moment decay, reached 3.389. At its tuned rate the control had a span of 0.107 averaged over the layers, below the span of AdamW at each of the five rates of its grid. At every higher rate, the span of the control exceeded all five. At $10^{-3}$, for example, the control reached a loss of 3.405, close to that of AdamW at its tuned rate, with a span of 0.211. Muon at $1.6 \times 10^{-2}$ and $3.2 \times 10^{-2}$, four and eight times the selected rate, had 42 and 37 loss spikes (defined in \cref{app:stress}) in 20 and 21 episodes of consecutive spikes at steps 1,096 to~3,513, each with the span of the first MoE layer at 0.992 to~1.001 at the nearest logged step, and no other run of the tuning of the small model had one.

\subsection{Stress Test}
\label{app:stress}

\begin{figure}[ht]
  \centering
  \includegraphics{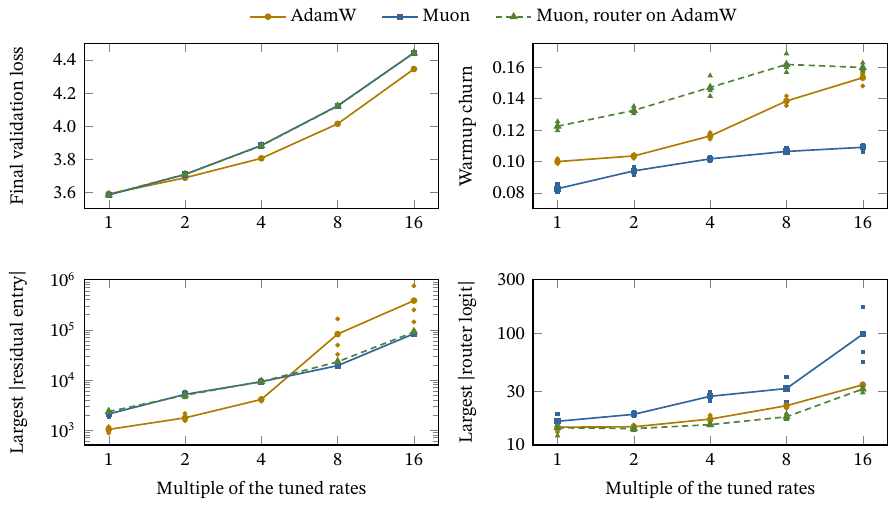}
  \caption{The stress test, on the small model with 0.5B tokens, against the multiple of the tuned rates. The panels show the final validation loss, the warmup churn, and the largest absolute entries of the residual stream and of the router logits over the run. Lines join the means over seeds 0 to~2, and in the last three panels the small marks are the seeds. The losses of the two Muon partitions coincide.}
  \label{fig:stress}
\end{figure}

\Cref{fig:stress} shows the runs of the stress test. The losses of the two Muon partitions rose more than that of AdamW above the tuned rates. Their learning-rate sensitivity, the mean over the multiples 1 to~8 of the final loss minus the lowest of these final losses, which follows \citet{wortsman2024proxies} over a narrower range of rates, was 0.24 against 0.18 for AdamW\@. Up to four times the tuned rates, the largest absolute entries of the residual stream under the Muon partitions were two to three times those of AdamW, in line with the larger activations under Muon that the report of Qwen3.8-Flash-Next found at twice the best rate~\citep{qiu2026qwennext}. From eight times, those of AdamW were larger, so the largest activations did not rank the partitions as the loss did. The largest absolute router logit stayed below 36 whenever AdamW trained the router, and under Muon its mean over the seeds grew from 16 at the tuned rates to~98 at 16 times, with 170 at one seed. A step is a spike when its training loss exceeds the median of the 201 steps centered on it by more than 0.1, the window length and threshold of Qwen3.8-Flash-Next~\citep{qiu2026qwennext}, and by more than five robust standard deviations of that difference~\citep{k2team2025k2v2}. Under this rule, the four experiments on the small model with 1B tokens had one spike, in a run of AdamW in the main comparison. The 45 runs at the five multiples had seven episodes of consecutive spikes. Six were at 16 times the tuned rates, three under Muon, two under AdamW, and one with the router on AdamW, and the seventh was under AdamW at twice the tuned rates. The three under Muon were all at one seed, each with the span of the first MoE layer within $2u$ of its ceiling, and at the spike with the router on AdamW this span was 0.946. With the fast bias at four times the tuned rates, Muon reached the same loss as with the usual bias and a lower warmup churn.

\subsection{Ablation and Its Probe}
\label{app:ablation}

\begin{figure}[t]
  \centering
  \includegraphics{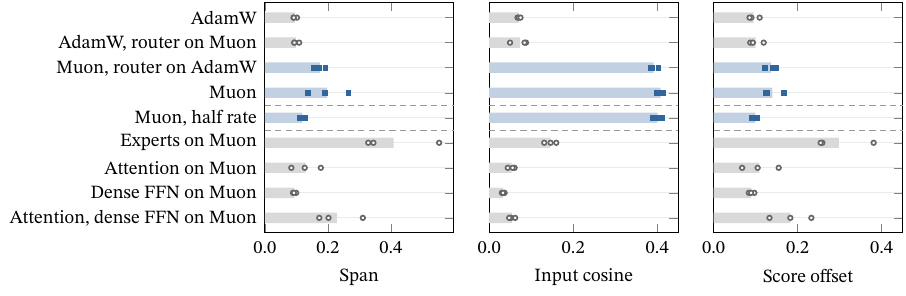}
  \caption{The span, the input cosine, and the score offset of the first MoE layer in the arms of the ablation with the bias, on the small model with 1B tokens. The bars are the means over seeds 3 to~5 and the marks are the seeds. Dashed lines separate the four combinations of the optimizers of the router and the backbone, Muon at half rate, and the four partitions with part of the backbone on Muon. The arms with the whole backbone on Muon are in blue.}
  \label{fig:arms}
\end{figure}

\begin{figure}[ht]
  \centering
  \includegraphics{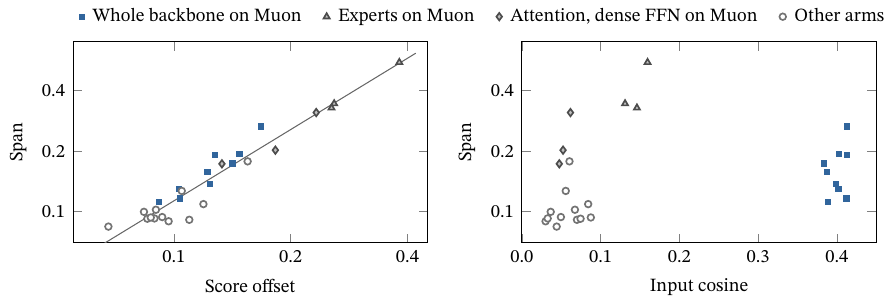}
  \caption{The span of the first MoE layer against its score offset (left, with the least-squares line of the logarithms) and against its input cosine (right), with one mark per run of the ablation with the bias. The span and the score offset are on logarithmic axes. Squares are the runs with the whole backbone on Muon, triangles those with only the routed experts on Muon, diamonds those with only the attention and the dense FFN on Muon, and circles the other four arms.}
  \label{fig:geometry}
\end{figure}

\Cref{fig:arms} shows the span, the input cosine, and the score offset of the first MoE layer in every arm of the ablation with the bias, and \cref{fig:geometry} the span against the other two. With only the routed experts on Muon, the span of the first MoE layer at seed~3 was 0.553, and about half of it came from one expert, since the span of the other 31 was 0.293. This expert received 2.5 times its fair share over steps 501 to~1,262 and had the lowest bias of the layer from step~1,263 to the end of the run.

Without the bias, the final usage perplexity averaged over the layers was 17.2 under Muon against 11.0 under AdamW, out of 32. In the first MoE layer it was 6.5 against 9.6, and under Muon the early overload of this layer was 7.7 to~7.9, close to the ceiling of 8. Both differences in the usage perplexity had the same sign at every seed.

In the probe of the ablation (\cref{sec:experiments}), we measured the two factors of the spread of the offsets in the first MoE layer over steps 501 to~3,000. The first, the length of the mean input divided by the average length of a single input, was 0.63 to~0.67 in every run with the backbone on Muon, whatever its rate, balancer, or router, and 0.21 in the run of AdamW\@. The second, the spread of the rows along the mean input, is the largest minus the smallest component along the direction of the mean input among the rows of the router, times the average length of a single input, so that at every step the product of the two factors is the spread of the offsets. Relative to the run of AdamW at seed~3, this spread was 3.0 under Muon and 1.4 at half the rate at the same seed. In the runs with the fast bias at seeds 3 to~5, it was 0.69 to~1.57 times that of this run, and in those with the router on AdamW and the backbone at half the rate, 3.6 to~3.7 times. The spread of the offsets under Muon, measured directly, was 9.8 times that of AdamW\@. With the fast bias, the early overload was 1.44 to~1.81, and over steps 501 to~3,000 the rise fell by 75\%.

\subsection{Large Model}
\label{app:large}

Over the last 10,000 steps, the busiest expert of the first MoE layer of the large model received 2.2 to~2.7 times its fair share under Muon and 2.3 to~2.8 times with the router on AdamW, against 1.07 under AdamW\@.

The runs with the router on AdamW had 2, 6, and 2 loss spikes at the three seeds, in 2, 4, and 1 episodes of consecutive spikes, at steps 1,010 to~2,349, and no run of the other two partitions had one. At nine of these ten spikes, the span of the first MoE layer at the nearest logged step was 1.000 or 1.001, about $2u$ or less below its ceiling of $1 + 2u$, and at the tenth it was 0.992. At each spike the busiest expert of this layer received 7.2 to~7.8 times its fair share. In the tuning of the large model, only Muon at $4 \times 10^{-3}$, twice the rate the tuning selected, had loss spikes, eight in seven episodes at steps 1,373 to~3,772, and at each of them the span of the first MoE layer at the nearest logged step was 1.000 or 1.001.

In the unchanged runs of the probe of the large model at seed~0 (\cref{sec:experiments}), the scores of the first MoE layer sank toward zero with the backbone on Muon, into the flat part of the sigmoid. With the router on AdamW, 84\% of them were below 0.01 over steps 2,001 to~3,000, against 28\% under Muon and none under AdamW\@. The bias then made most of the choices. Of the selected experts, 80\% with the router on AdamW and 53\% under Muon would not have been selected by the scores alone, against 26\% under AdamW\@. Centering kept the scores off the flat part of the sigmoid under both Muon partitions at every seed. The early overload, over steps 501 to~2,788, fell to~1.11 to~1.15 times the fair share, and over steps 501 to~3,000 the span was 0.19 to~0.23, below the 0.28 to~0.38 of AdamW over the same steps of its runs at seeds 0 to~2. Centering changed the loss at step~3,000 by $-0.0021$ to~$+0.0001$. With the fast bias at seed~0, the early overload was 2.29 against 5.59 with the usual bias, and over steps 501 to~3,000 the span was 0.375 against 0.716 and the load churn of the layer 0.083, below the 0.135 of the usual bias but three times the 0.026 of AdamW\@.

\section{Reproduction}
\label{app:reproduction}

{\setlength{\rightskip}{0pt plus 5em}\spaceskip=\fontdimen2\font\relax The code, the Python package \texttt{tfg} with its 213 tests, is public at \url{https://github.com/marcoshernanz/muon-moe-routing}, with the commands that train and analyze every run in \file{code/RUNBOOK.md}. The tokenized training data are the public Hugging Face dataset \href{https://huggingface.co/datasets/marcoshernanz/llm-lab-fineweb-edu-sample10bt-bpe-16384-full}{\nolinkurl{marcoshernanz/llm-lab-fineweb-edu-sample10bt-bpe-16384-full}}, and the directory of every run, with its logs and analysis, is in \href{https://huggingface.co/datasets/marcoshernanz/tfg-runs}{\nolinkurl{marcoshernanz/tfg-runs}}, where \file{claims.csv} in the analysis of the main comparison and of the ablation, bias-speed, balance-loss, and large-model experiments lists their planned comparisons with an expected sign, with their outcomes, and some comparisons that the plan set apart. The balance-loss file repeats seven of the planned comparisons of the ablation. The planned comparisons of the stress test and those without an expected sign, with their outcomes, are in \cref{app:program}. The figures of this paper are made from the runs by \file{code/ai-experiments/thesis_figures_2026_09/extract.py} and \file{code/ai-experiments/paper_figures_2026_10/extract.py}, and its tables by \file{code/ai-experiments/paper_figures_2026_10/tables.py}, which reads the analysis of each experiment from the run dataset.\par}

\end{document}